\def\year{2017}\relax
\documentclass[letterpaper]{article} 
\usepackage{aaai17}  
\usepackage{times}  
\usepackage{helvet}  
\usepackage{courier}  
\usepackage{url}  
\usepackage{graphicx}  

\usepackage{pdfpages}
\usepackage{subcaption}
\usepackage{multirow}

\usepackage{amsmath}
\usepackage{amsfonts}
\usepackage{amssymb}

\usepackage{xspace}
\newcommand*{\eg}{e.g.\@\xspace}
\newcommand*{\ie}{i.e.\@\xspace}

\begin{document}
\title{Cultural Diffusion and Trends in Facebook Photographs}
\author{
Quenzeng You  \\
University of Rochester \\
qyou@cs.rochester.edu \\
\And
Dar{\'\i}o Garc{\'\i}a-Garc{\'\i}a \\ 
Facebook \\
dariogg@fb.com\\
\And 
Manohar Paluri \\ 
Facebook \\
mano@fb.com\\
\And 
Jiebo Luo \\ 
University of Rochester \\
jluo@cs.rochester.edu\\
\And 
Jungseock Joo\thanks{Corresponding author (jjoo@commstds.ucla.edu).  This research was mainly performed while Q. You was a research intern and J. Joo was a research scientist at Facebook.
} \\
UCLA \\
jjoo@commstds.ucla.edu\\
}

\maketitle
\begin{abstract}
Online social media is a social vehicle in which people share various moments of their lives with their friends, such as playing sports, cooking dinner or just taking a selfie for fun, via visual means, i.e., photographs. Our study takes a closer look at the popular visual concepts illustrating various cultural lifestyles from aggregated, de-identified photographs. We perform analysis both at macroscopic and microscopic levels, to gain novel insights about global and local visual trends as well as the dynamics of interpersonal cultural exchange and diffusion among Facebook friends. We processed images by automatically classifying the visual content by a convolutional neural network (CNN). Through various statistical tests, we find that socially tied individuals more likely post images showing similar cultural lifestyles. To further identify the main cause of the observed social correlation, we use the Shuffle test and the Preference-based Matched Estimation (PME) test to distinguish the effects of influence and homophily. The results indicate that the visual content of each user's photographs are temporally, although not necessarily causally, correlated with the photographs of their friends, which may suggest the effect of influence. Our paper demonstrates that Facebook photographs exhibit diverse cultural lifestyles and preferences and that the social interaction mediated through the visual channel in social media can be an effective mechanism for cultural diffusion.
\end{abstract}

\section{Introduction}
Online social networks allow people to share news about their lives with friends, such as hobbies, vacations, events, their favorite foods or sports. This reflects the preferred lifestyles of individual users and collectively forms the ``culture'' of a society when there are commonly shared preferences by the members of a society.\footnote{The definition of culture often goes beyond physically apparent activities and also includes social values or beliefs \cite{kroeber1952culture}, but in this paper we focus on human activities common in our daily lives.}

Such lifestyles, \eg, what we eat, what we wear, or what we do, are important and popular topics of user generated content in social media, especially in user \textbf{photographs}. Many people take photographs about their daily activities and events with their smartphones and post online to share with friends. The popularity of online visual sharing has greatly surged in recent years with a rapid growth of or shift to visual-centric online media. Therefore, by analyzing the photographs people post and their content, we will be able to tell their preferences on certain lifestyles and also understand how popular lifestyles evolve over space and time. 

The primal goal of our paper is to understand the role of social media in the process of ``culture sharing'' which means the exchanges or mutual exposures of preferred lifestyles via social ties between users from different cultural backgrounds. For example, many users on Facebook have friends in other countries, who would post about their own local cuisines. The users will see these posts and photographs and may become interested in trying it. They can also make their own posts about their experience, which will be visible to their friends. This process is known as social influence. 

How can we examine the flow of cultural preference from user posts? This problem is closely related to the topic of information or behavior diffusion in social networks \cite{gruhl2004information,adar2005tracking,cha2009measurement,bakshy2012role}. These studies take advantage of shared network links or urls from different users as references or infer topics from text data. However, we are interested in detecting and comparing general lifestyles and preferences which are not specified by users but non-verbally depicted in the photographs, which necessitates \textit{visual content} analysis.

To this end, our paper makes two main contributions by a scalable computer vision pipeline. Firstly, we study cultural trends in Facebook photographs from 2013 to 2016. To protect user privacy, all photos analyzed were deidentified and aggregated. We first define commonly observable visual concepts related to lifestyles to quantify content of photographs. With these categories, we automatically classify the photographs by a convolutional neural network (CNN). We present various dynamic trends of different cultural lifestyles and activities, which show seasonal, geographical, or global trends.

Secondly, we also investigate the role of the friendship network in cultural diffusion in user photographs. People from diverse cultural backgrounds use Facebook to connect with friends and family. Social interactions within social media such as an exposure to a friend's photograph may make the user more likely to adopt a new preference and post similar photographs. In contrast to existing studies on behavior or information diffusion which rely on shared links or explicit annotations, we automatically classify visual content of photographs and compare the predicted scores. We first measure the social correlation of cultural lifestyles between friends. Then we further use advanced statistical tests to compare the effects between influence and homophily on the observed social correlation.

We summarize our key research questions as follows:
\begin{itemize}
\item Do Facebook photographs reflect the cultural preferences on lifestyles in different places and times?
\item Social Correlation: Do friends in Facebook post more similar photographs than non-friends?
\item Social Influence: Is the correlation, if any, due to homophily or influence?
\end{itemize}

\section{Related Work}
Recent studies in computer vision have analyzed visual content from social media or web data but without considering network structures or content flow and diffusion. Likewise, studies in data mining or network analysis typically do not employ visual content analysis at massive scale. Our study bridges the gap between two areas of research. 

\textbf{Visual Recognition for Web and Social Media.} 
Automated visual content analysis by computer vision has been used to analyze web images in various applications including fashion studies \cite{simo2015neuroaesthetics} or political analysis \cite{joo2014visual}. Geo-tagged photographs are particularly useful to understand local communities and geographic differences in popular photographic style and content \cite{redi2016makes}, architectural style \cite{doersch2012makes}, natural environment \cite{wang2013observing}, ecological phenomena \cite{zhang2012mining}, or other socio-economic statues \cite{zhou2014recognizing,salesses2013collaborative,ordonez2014learning,souza2015dawn}. 
These studies collect images for each geographical region and treat them collectively without distinguishing who post them (\ie, photos are used solely to study geographical features). In contrast, the key concern in this paper is each individual user's social relation, and we study the role of social ties in cultural diffusion.

A few studies have also examined how image features can predict image popularity \cite{khosla2014makes,totti2014impact} or viewer engagement \cite{bakhshi2014faces}. These studies focus on image instance-level analysis (\ie, ``what makes \textit{this} image popular?'') whereas our paper investigates whether certain visual concepts propagate between images of different users.

\textbf{Diffusion in Social Networks.}
Social correlation and behavioral diffusion in social networks is another active research topic. Many studies have reported that behaviors or preferences of people can spread via social ties in social networks \cite{bond201261,lewis2012social,christakis2013social,aral2011creating}. For example, a longitudinal study \cite{lewis2008tastes} showed that online friends tend to share similar cultural tastes on movie, music, or books, but a subsequent analysis \cite{lewis2012social} revealed that these tastes are rarely contagious. These studies exploited user surveys or other attributes declared by users (\eg, profile information). In contrast, we infer the latent cultural preferences from user photographs.

Previous research has commonly identified three underlying factors driving social correlations: homophily \cite{mcpherson2001birds}, influence \cite{rogers2010diffusion}, and confounding factors. To distinguish the effects between homophily and influence from observational data, a handful of statistical tests have been proposed. Among them, we adopt the Shuffle test \cite{anagnostopoulos2008influence} and the PME test \cite{sharma2016distinguishing} because these are highly scalable to our large scale data than other methods (\eg, a simulation from a full joint state distribution \cite{snijders2010introduction}).

\begin{table}
\centering

\begin{tabular}{|p{1.6cm}|p{5.5cm}|}
\hline
Category & Concepts \\
\hline
\hline
Sports & baseball, basketball, climbing, football, golf, ski, soccer, swimming, tennis \ldots\\
\hline
Animals & bear, bird, bug, cat, cow, crocodile, deer, dog, horse, spider, tiger \ldots\\
\hline
Clothes & backpack, bikini, boots, dress, hat, heels, sunglasses, ties, \ldots\\
\hline
Food & avocado, bagel, banana, beer, blueberry, icecream, pizza, salad, sushi \ldots\\
\hline
Furniture  & bookshelf, bed, chair, kitchen, table,\ldots\\
\hline
Music & accordion, cello, flute, guitar, piano,\ldots\\
\hline
Plants & flower, grass, trees, bush, \ldots\\
\hline
Structures & bridge, house, chimney, monument, skyscraper, \ldots\\
\hline
Places & Big Ben, Colosseum, Eiffel tower, Louvre, Opera House \ldots\\
\hline
Scene & beach, closeup, fireworks, nature, night, selfie, sky, sunset, water,\ldots\\
\hline
Vehicles & bicycle, boat, bus, car, train, \ldots\\
\hline
\end{tabular}
\caption{A partial visual concept list in our analysis.}
\label{tab:concept}
\end{table}

\section{Data}
For our analysis, we use two groups of anonymized Facebook photographs. We did not use any user identifiable information. The social graph, friendship connections, was used in an aggregated form. The first set contains around 750 million de-identified photographs, sampled from the whole world in 2013-2016. Each photograph is associated with the photo upload time and the location of the owner. 

The second set contains around 250 million de-identified photographs, sampled from 1.3 million users living in the same location. Since social correlation can be caused by confounding factors such as user attributes, we control for user gender, age group, and location (at city level) in this dataset. Such a treatment will not completely rule out all possible alternative explanations. However, we separate and isolate each user group by user attributes to minimize the effects from these confounding variables. We use a complete network of users in this area so that we can examine the photograph similarity between people who are friends as well as who are not friends to each other. We chose Seattle metropolitan area because it has a reasonably large but tractable number of users.

\begin{figure*}
\centering
 \begin{subfigure}[b]{.23\textwidth}
 \includegraphics[width=\textwidth]{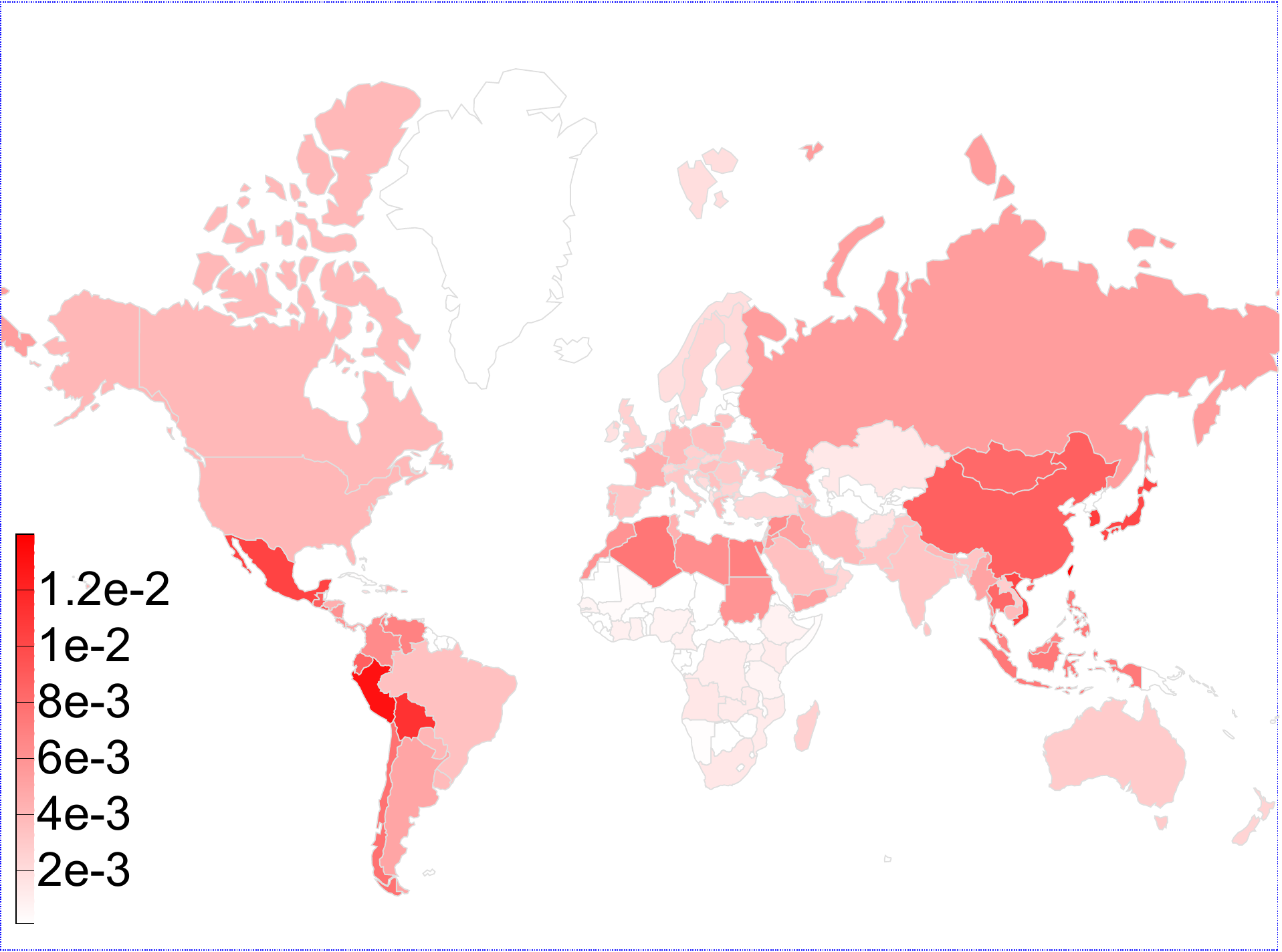}
 \caption{Baseball}
 \end{subfigure}
  \begin{subfigure}[b]{.23\textwidth}
 \includegraphics[width=\textwidth]{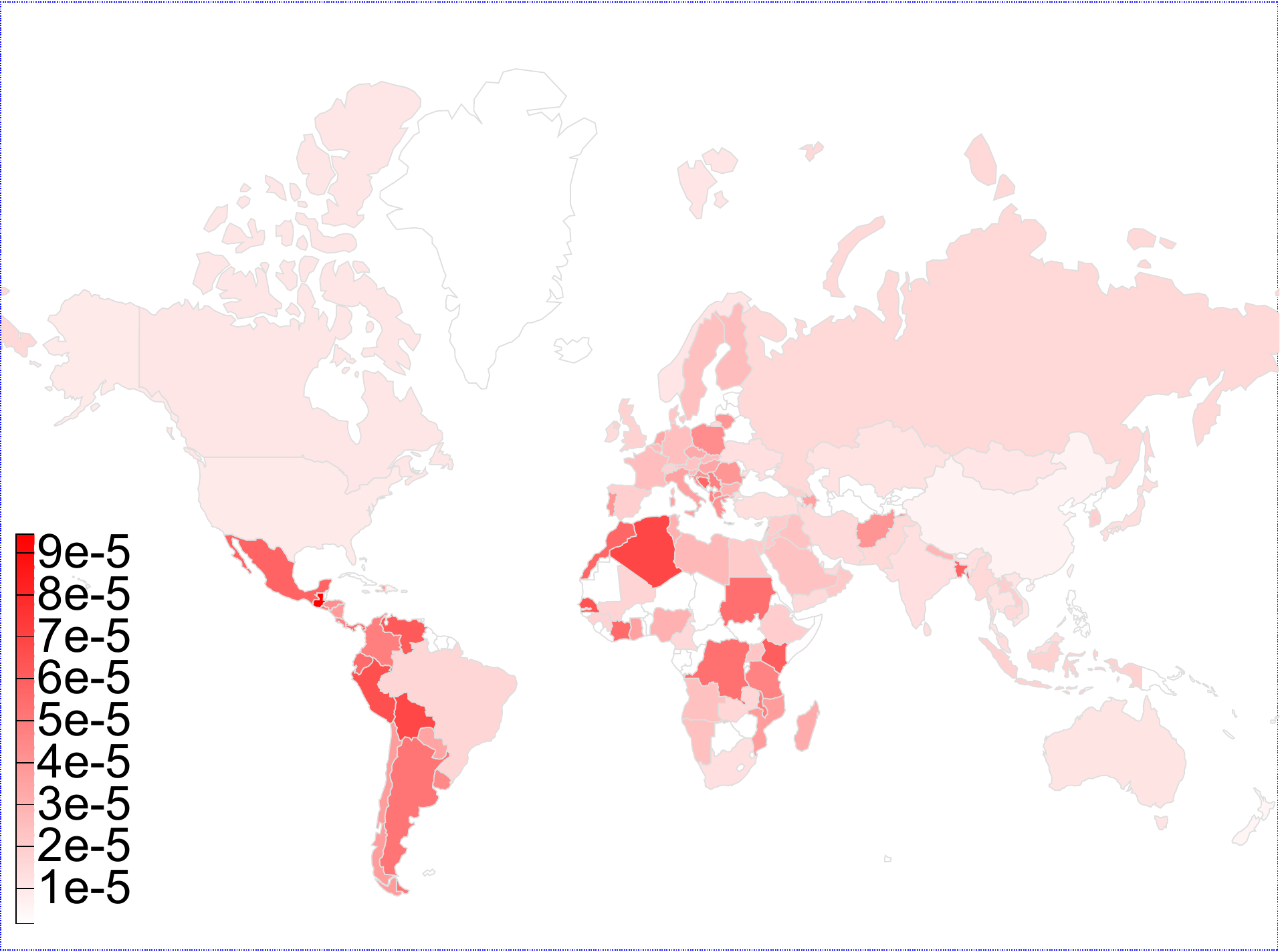}
 \caption{Soccer}
 \end{subfigure}
  \begin{subfigure}[b]{.23\textwidth}
 \includegraphics[width=\textwidth]{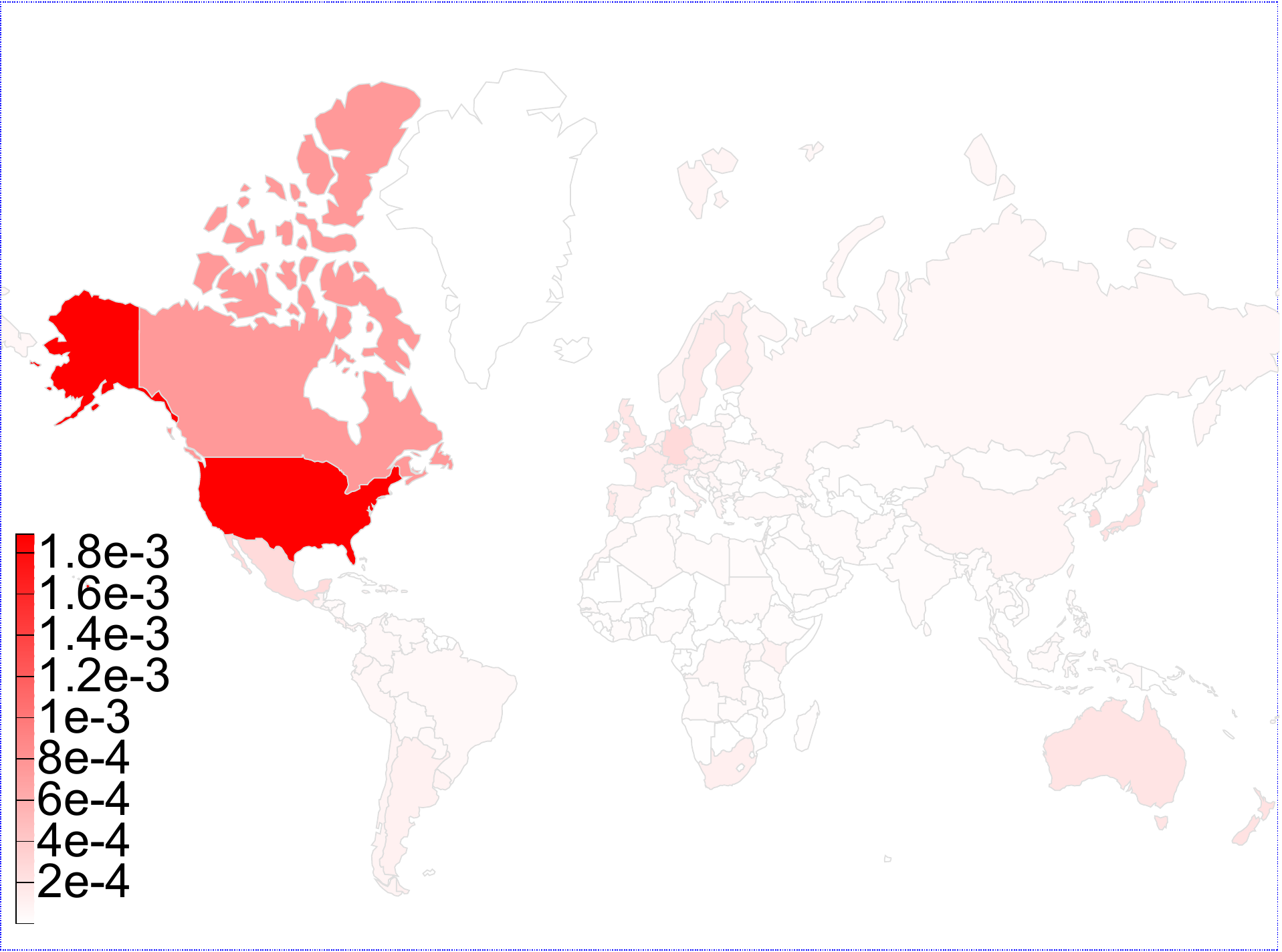}
 \caption{American Football}
 \end{subfigure}
  \begin{subfigure}[b]{.23\textwidth}
 \includegraphics[width=\textwidth]{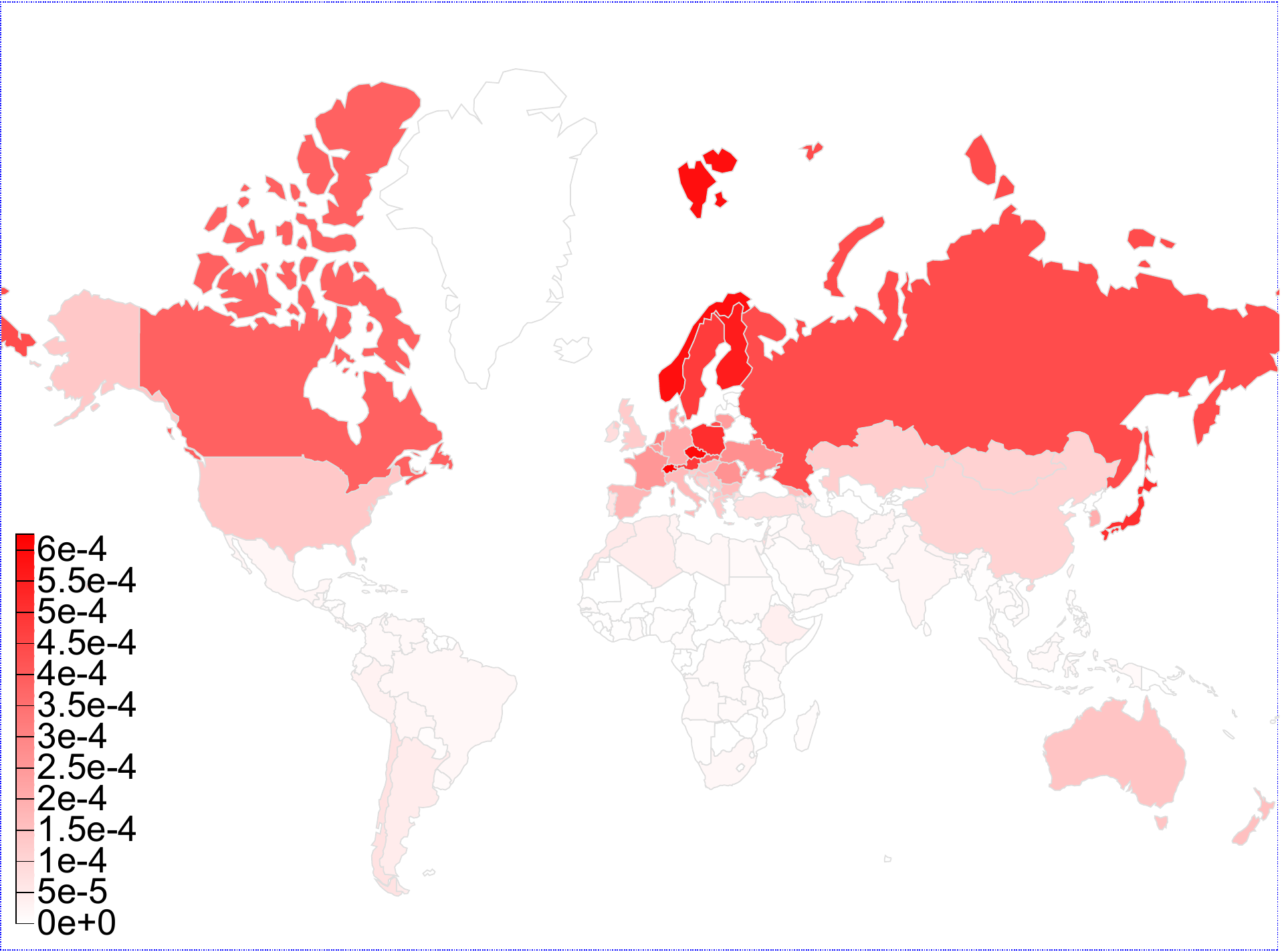}
 \caption{Snowboarding}
 \end{subfigure}
   \begin{subfigure}[b]{.23\textwidth}
 \includegraphics[width=\textwidth]{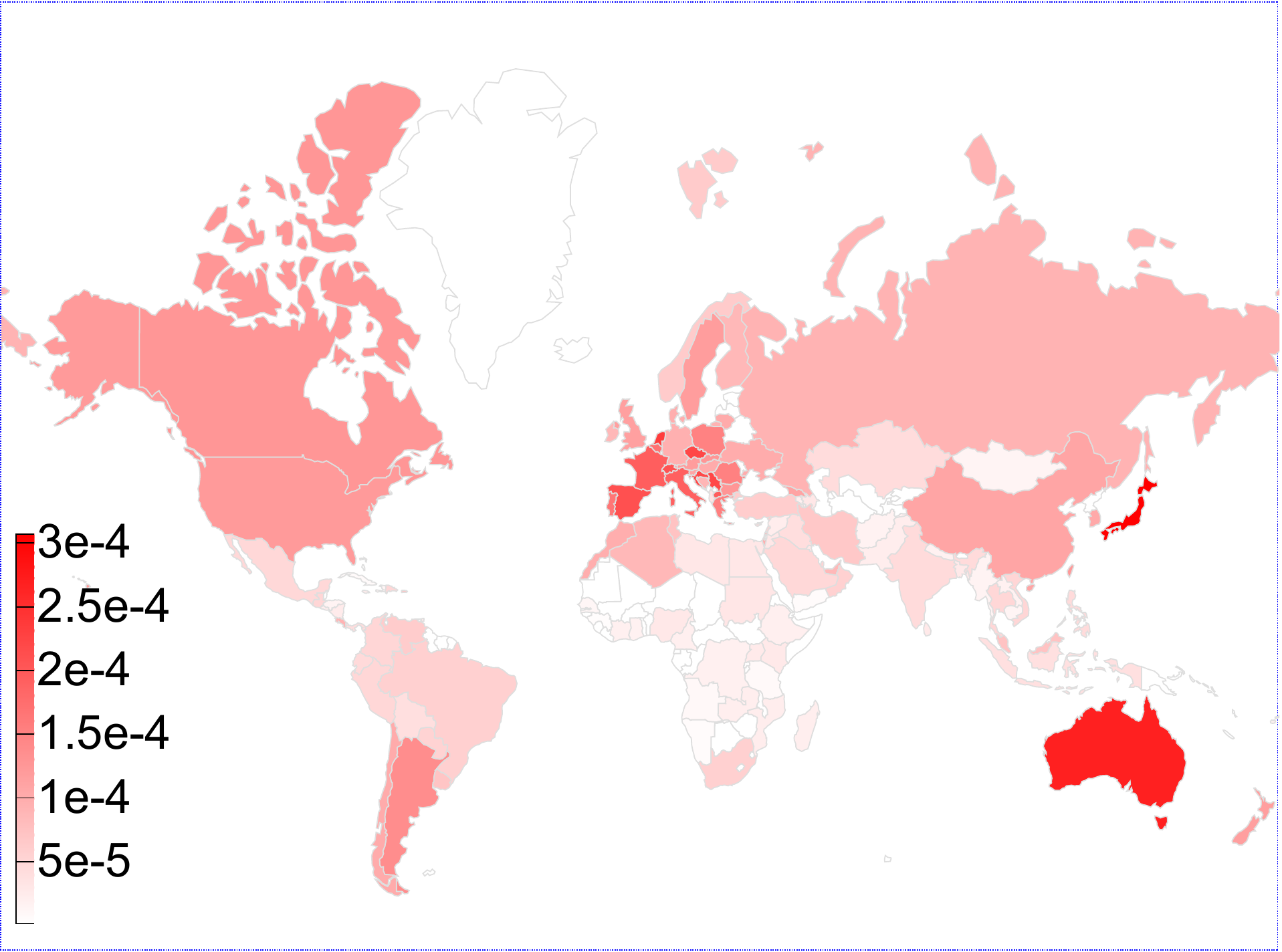}
 \caption{Tennis}
 \end{subfigure}
 \begin{subfigure}[b]{.23\textwidth}
 \includegraphics[width=\textwidth]{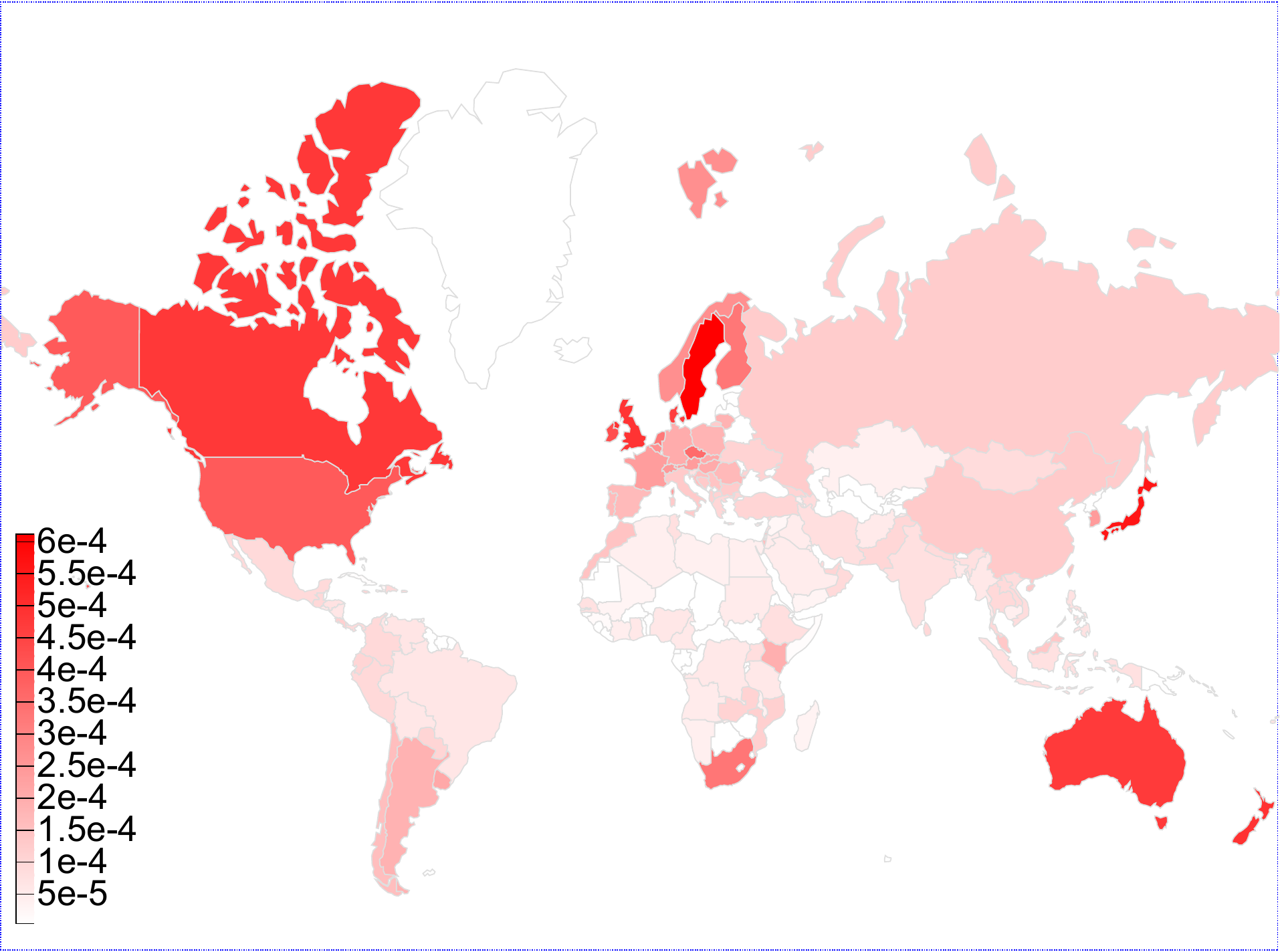}
 \caption{Golf}
 \end{subfigure}
  \begin{subfigure}[b]{.23\textwidth}
 \includegraphics[width=\textwidth]{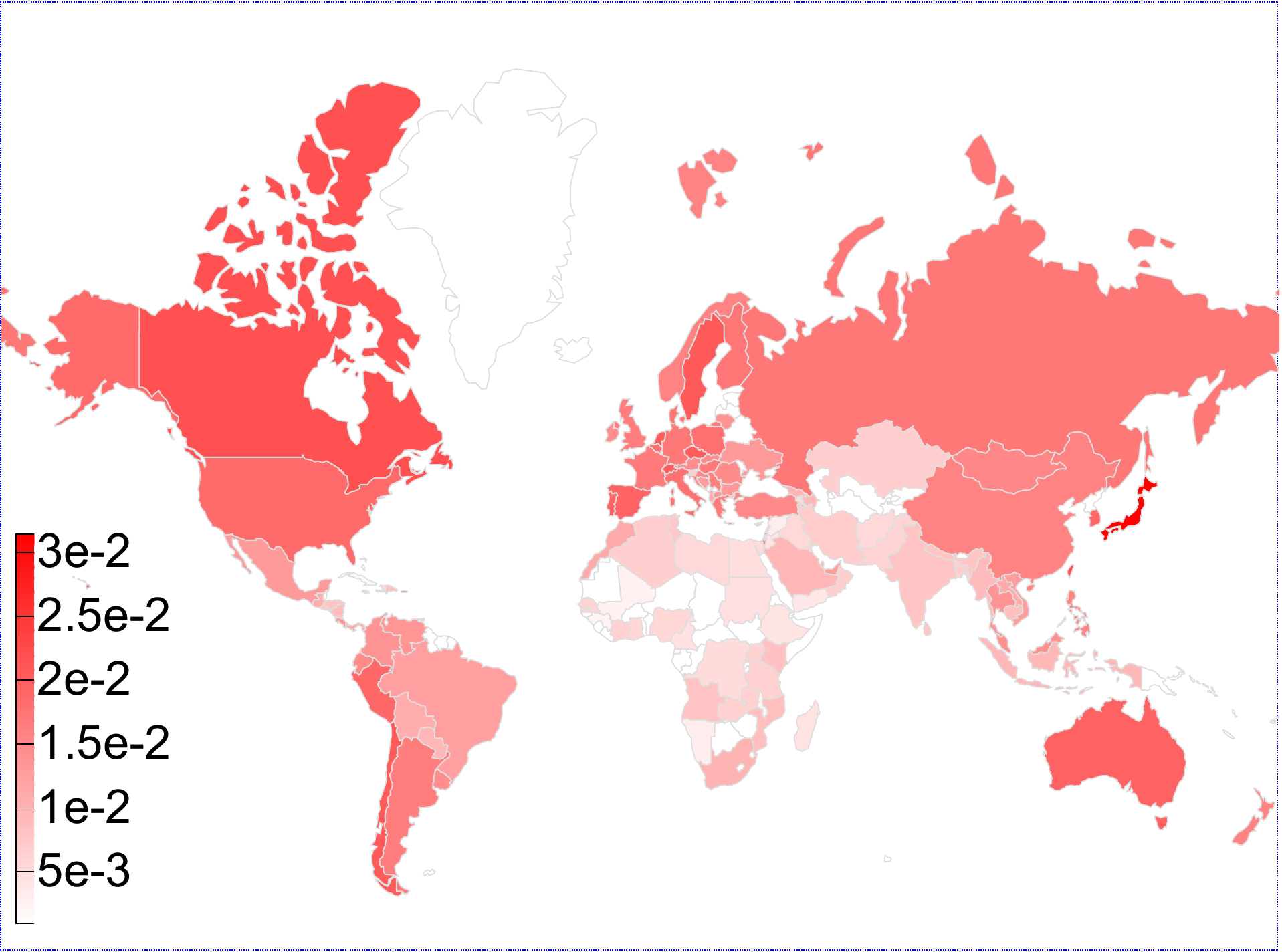}
 \caption{Basketball}
 \end{subfigure}
   \begin{subfigure}[b]{.23\textwidth}
 \includegraphics[width=\textwidth]{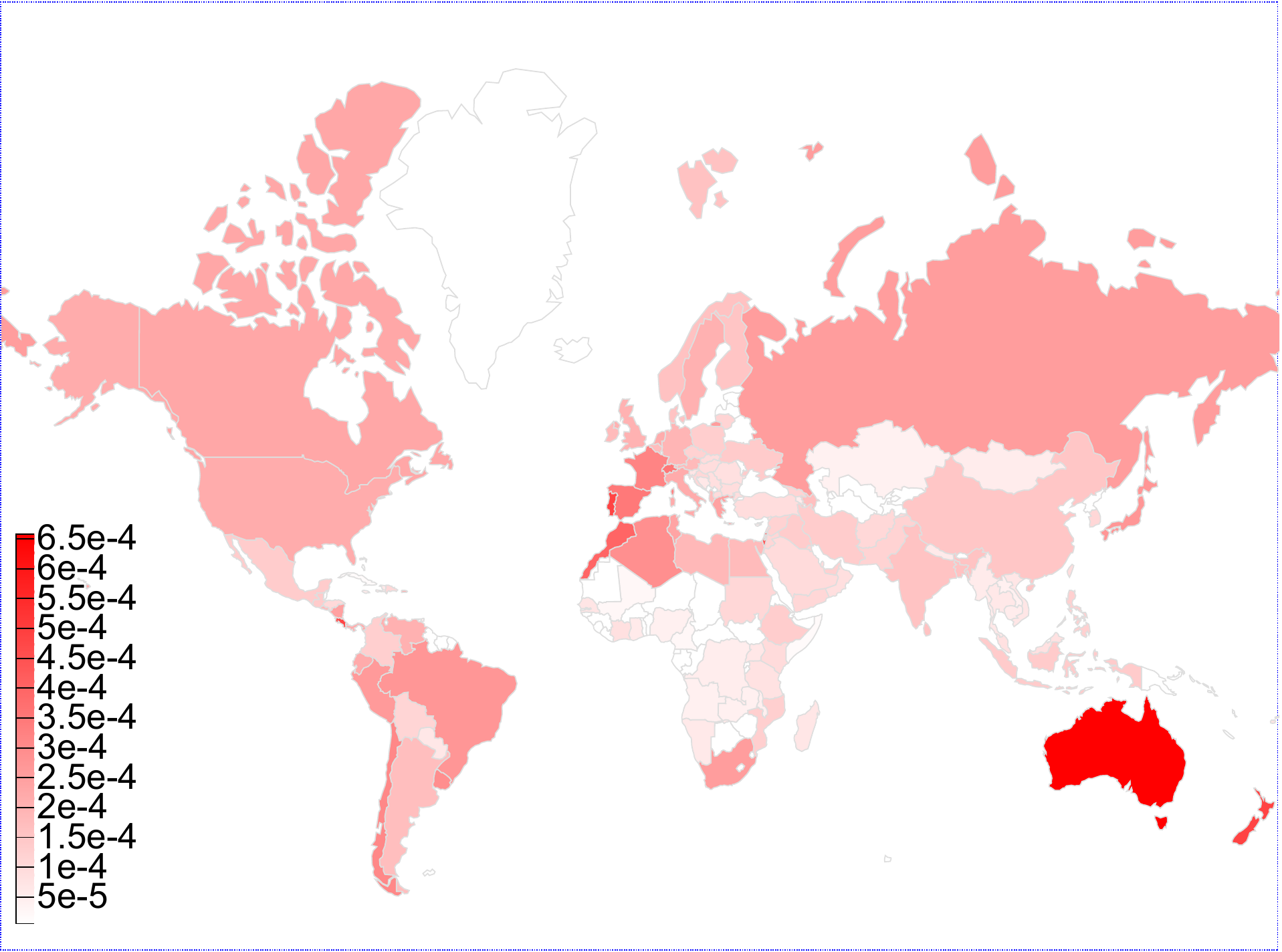}
 \caption{Surfing}
 \end{subfigure}
    \begin{subfigure}[b]{.23\textwidth}
 \includegraphics[width=\textwidth]{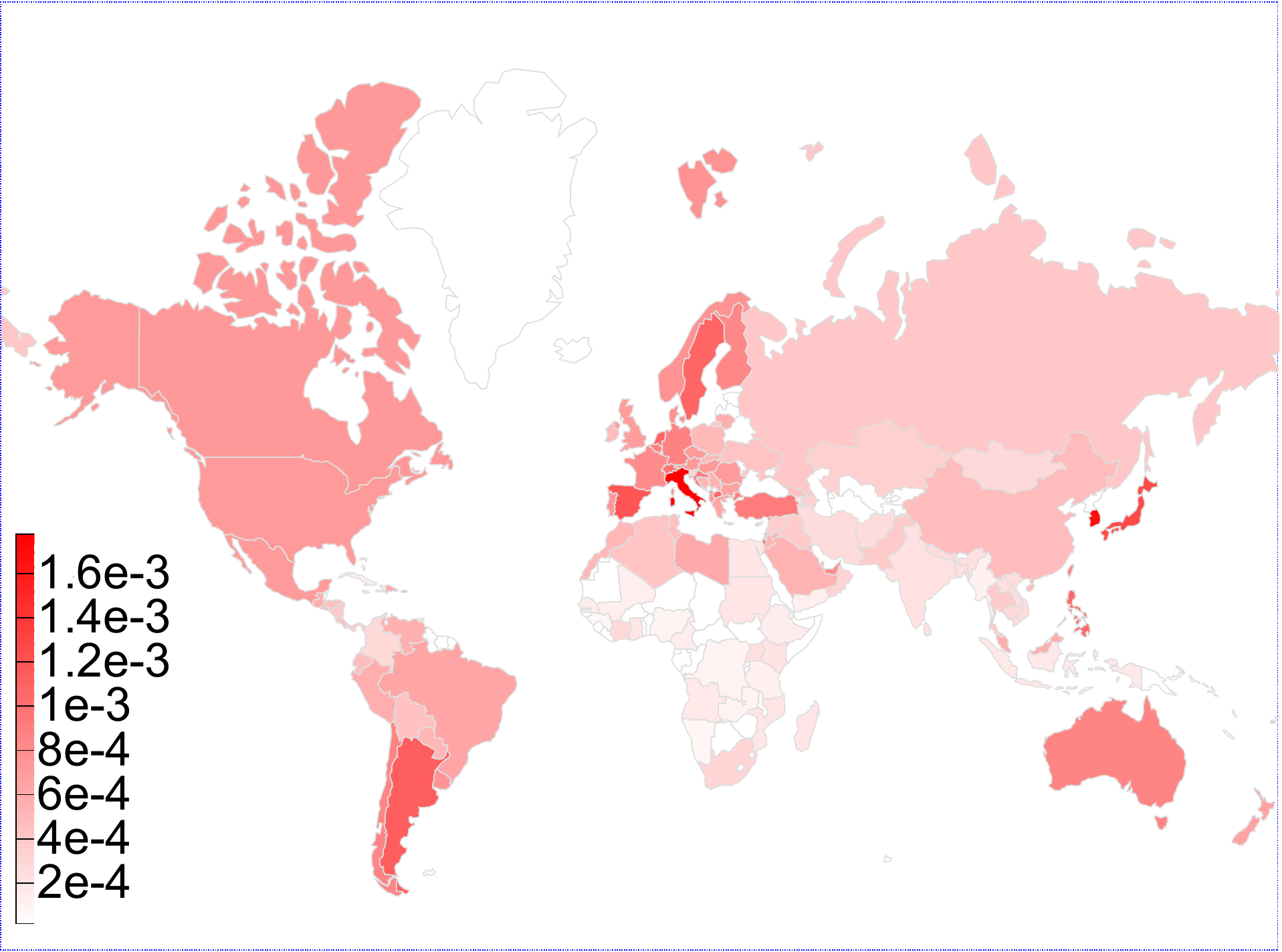}
 \caption{Pizza}
 \end{subfigure}
    \begin{subfigure}[b]{.23\textwidth}
 \includegraphics[width=\textwidth]{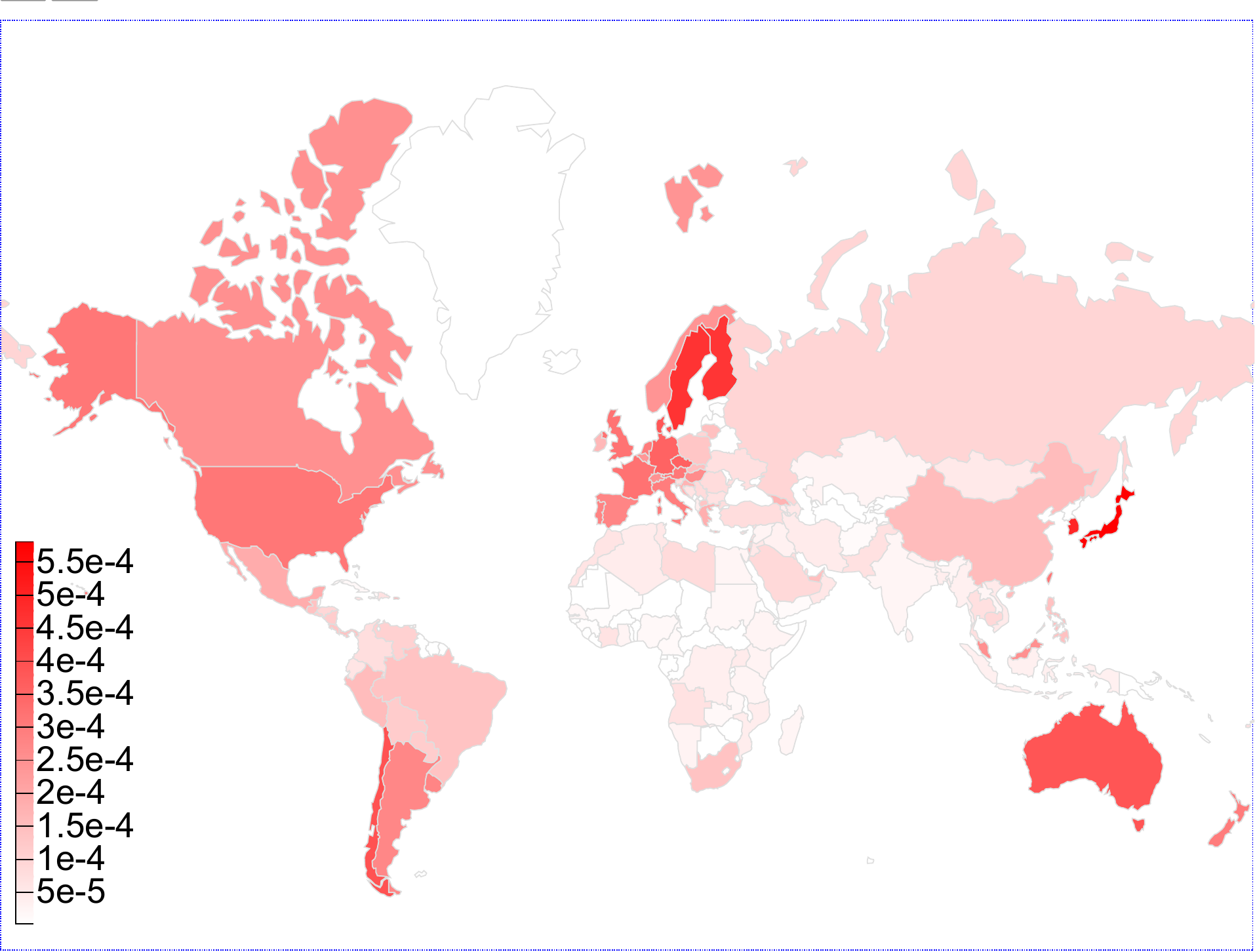}
 \caption{Hamburger}
 \end{subfigure}
    \begin{subfigure}[b]{.23\textwidth}
 \includegraphics[width=\textwidth]{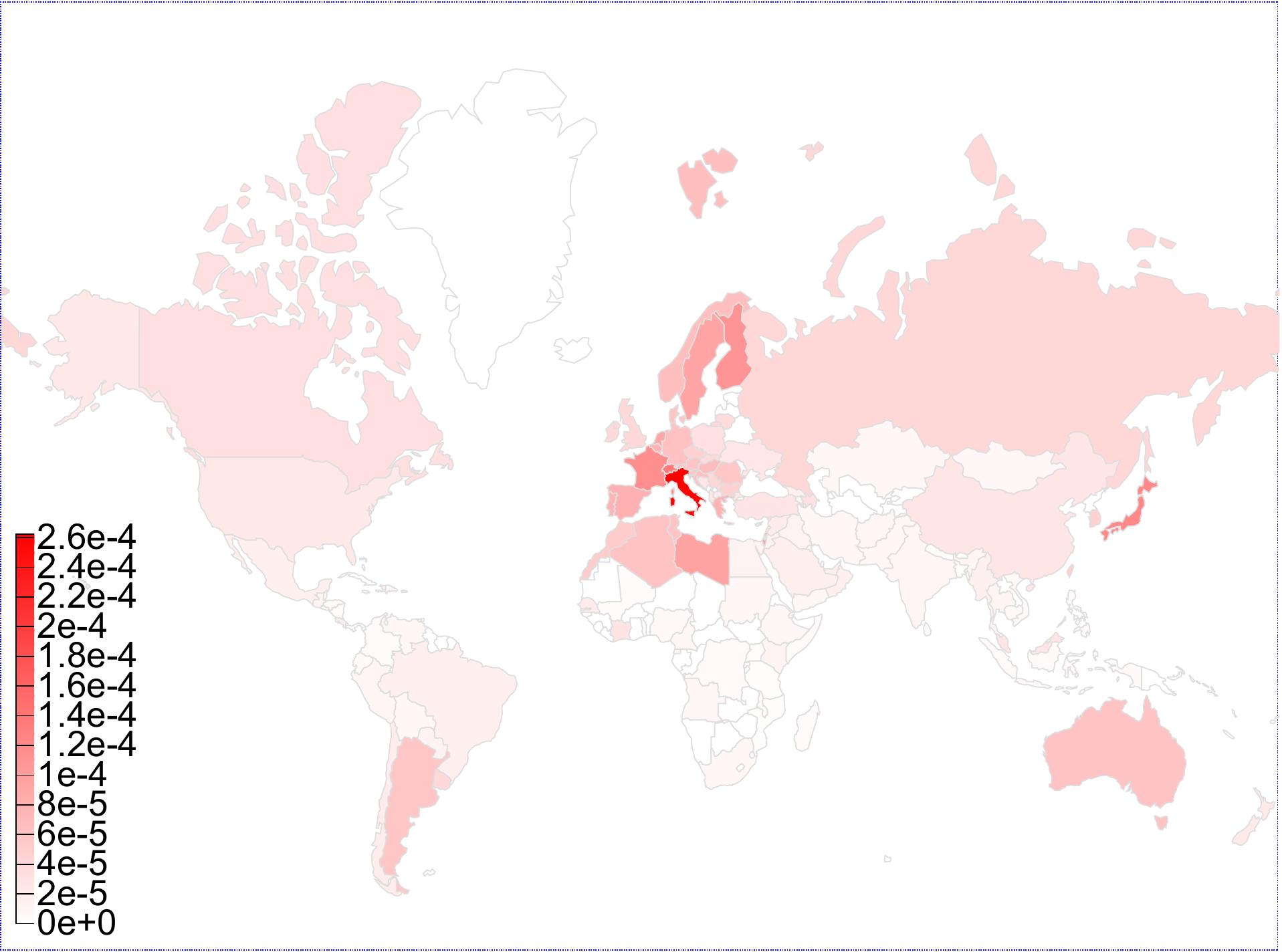}
 \caption{Croissant}
 \end{subfigure}
    \begin{subfigure}[b]{.23\textwidth}
 \includegraphics[width=\textwidth]{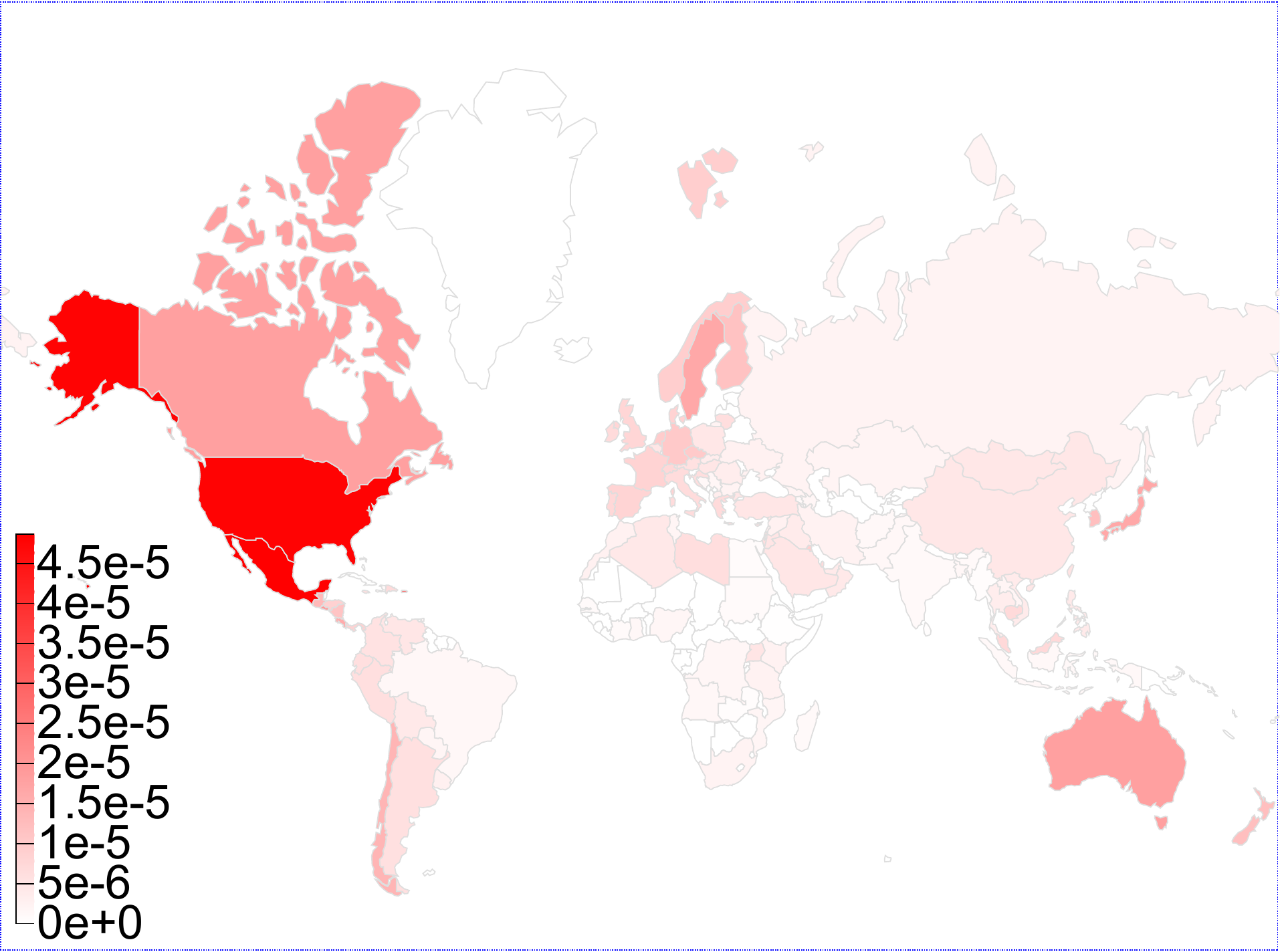}
 \caption{Tacos}
 \end{subfigure}
    \begin{subfigure}[b]{.23\textwidth}
 \includegraphics[width=\textwidth]{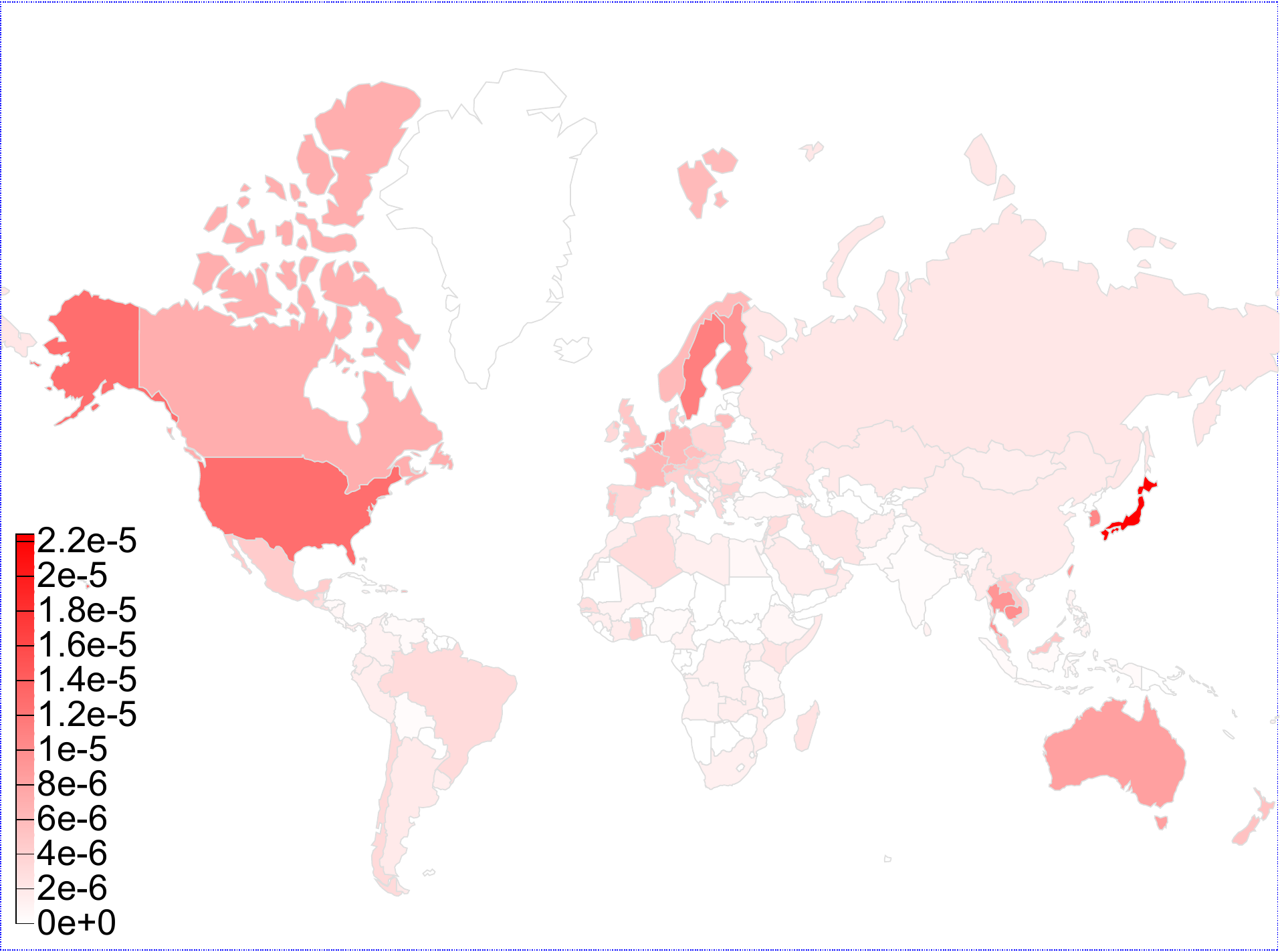}
 \caption{Salad}
 \end{subfigure}
    \begin{subfigure}[b]{.23\textwidth}
 \includegraphics[width=\textwidth]{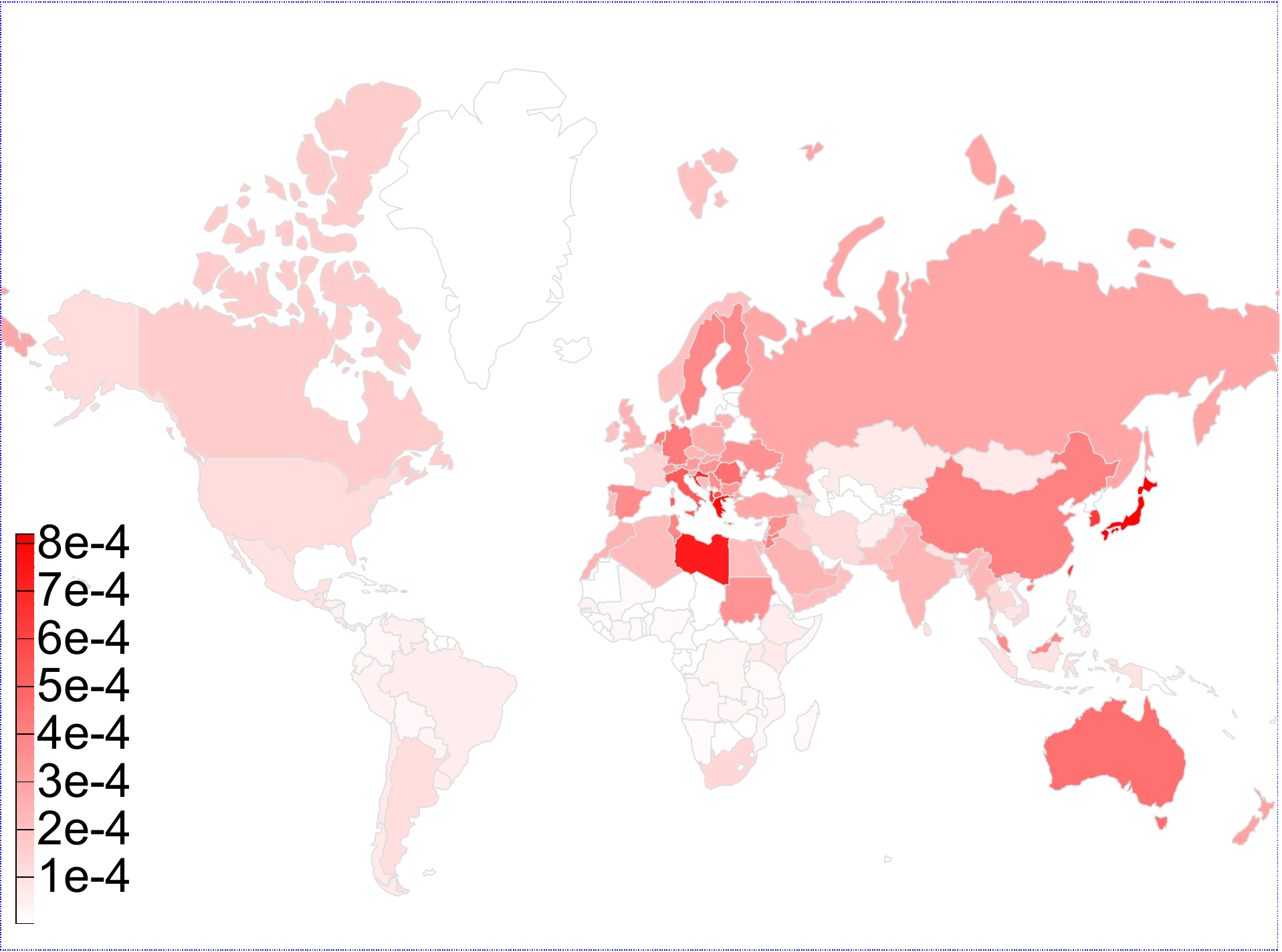}
 \caption{Latte}
 \end{subfigure}
    \begin{subfigure}[b]{.23\textwidth}
 \includegraphics[width=\textwidth]{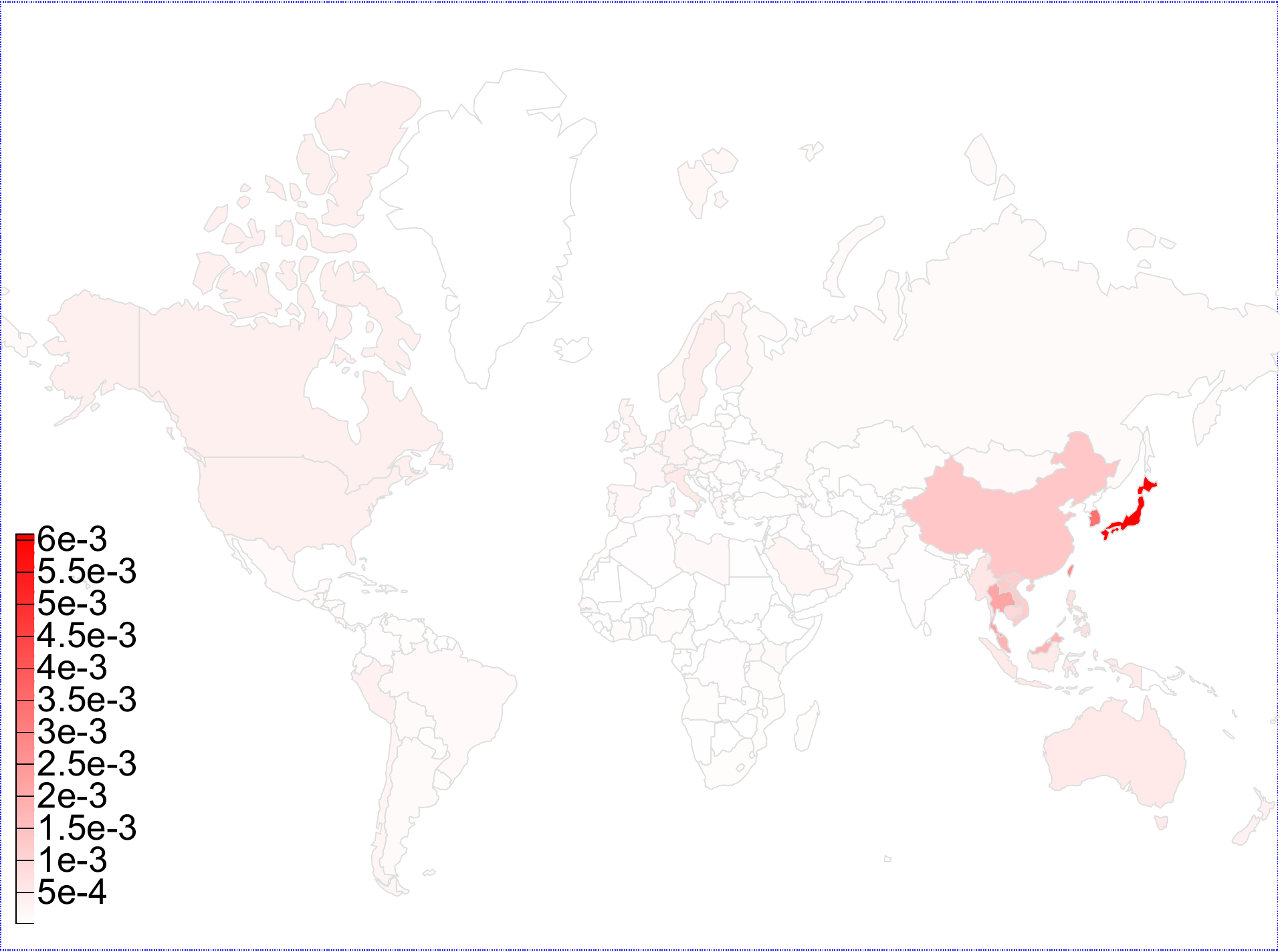}
 \caption{Noodle}
 \end{subfigure}
    \begin{subfigure}[b]{.23\textwidth}
 \includegraphics[width=\textwidth]{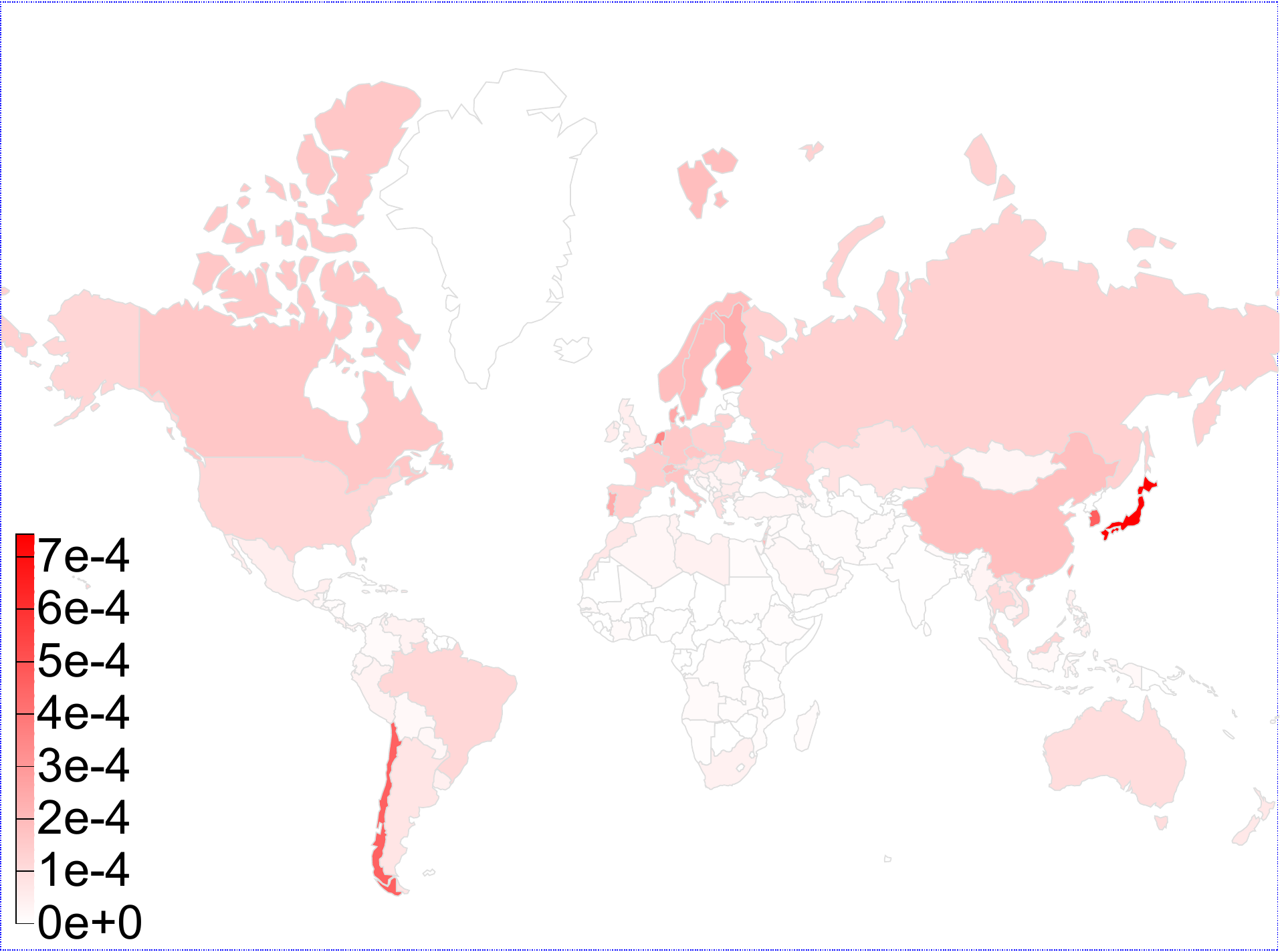}
 \caption{Sushi}
 \end{subfigure}
\caption{Popular concepts in the sports and food categories across different countries, aggregated from July 2013 to June 2016. More red means a higher average score of photographs in that country for the concept. See the text for detail. }
\label{fig:global}
\end{figure*}

\begin{figure*}[!htbp]
\centering
\begin{subfigure}[b]{.32\textwidth}
 \includegraphics[width=\textwidth]{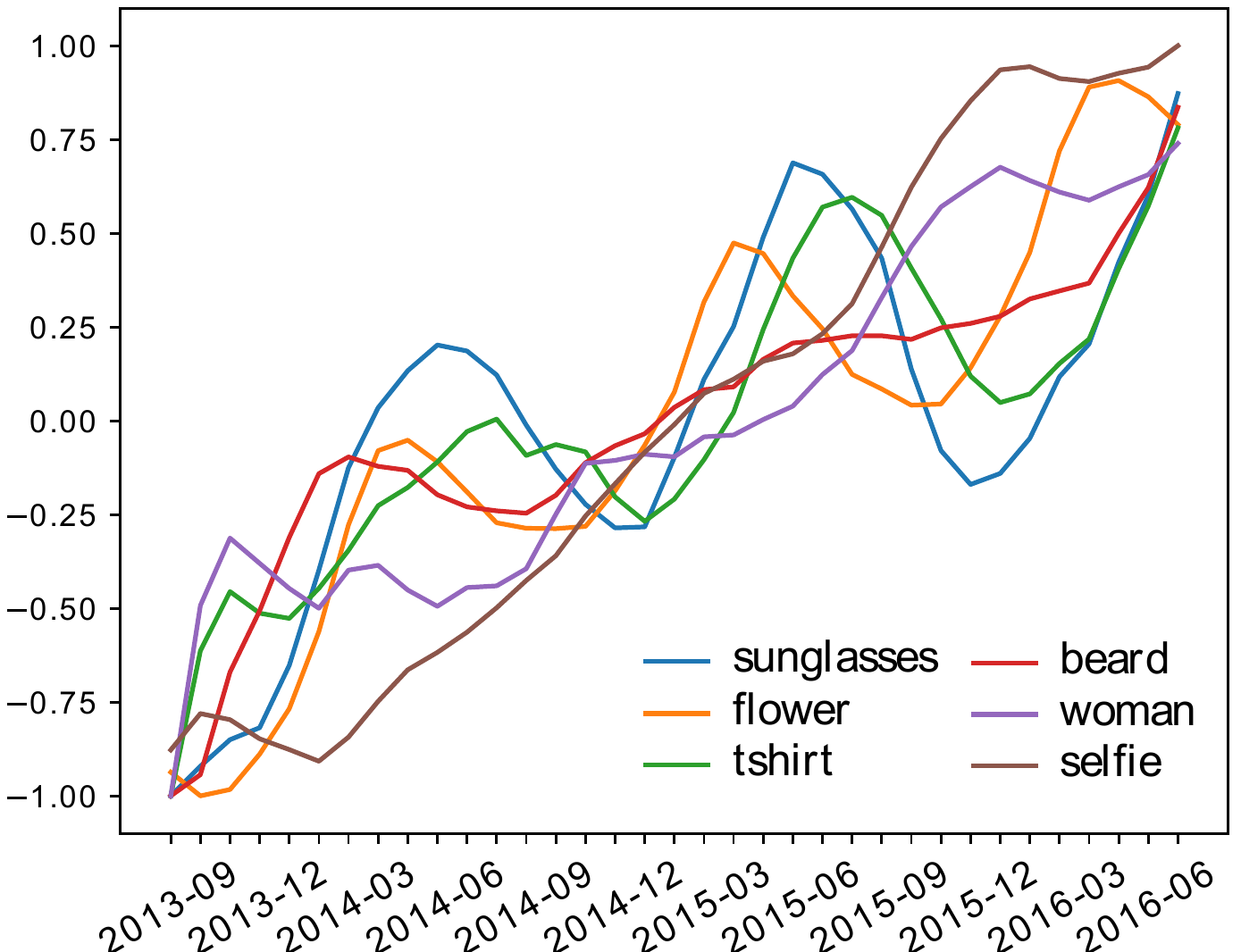}
 \caption{Increasing}
 \end{subfigure}
 \begin{subfigure}[b]{.32\textwidth}
 \includegraphics[width=\textwidth]{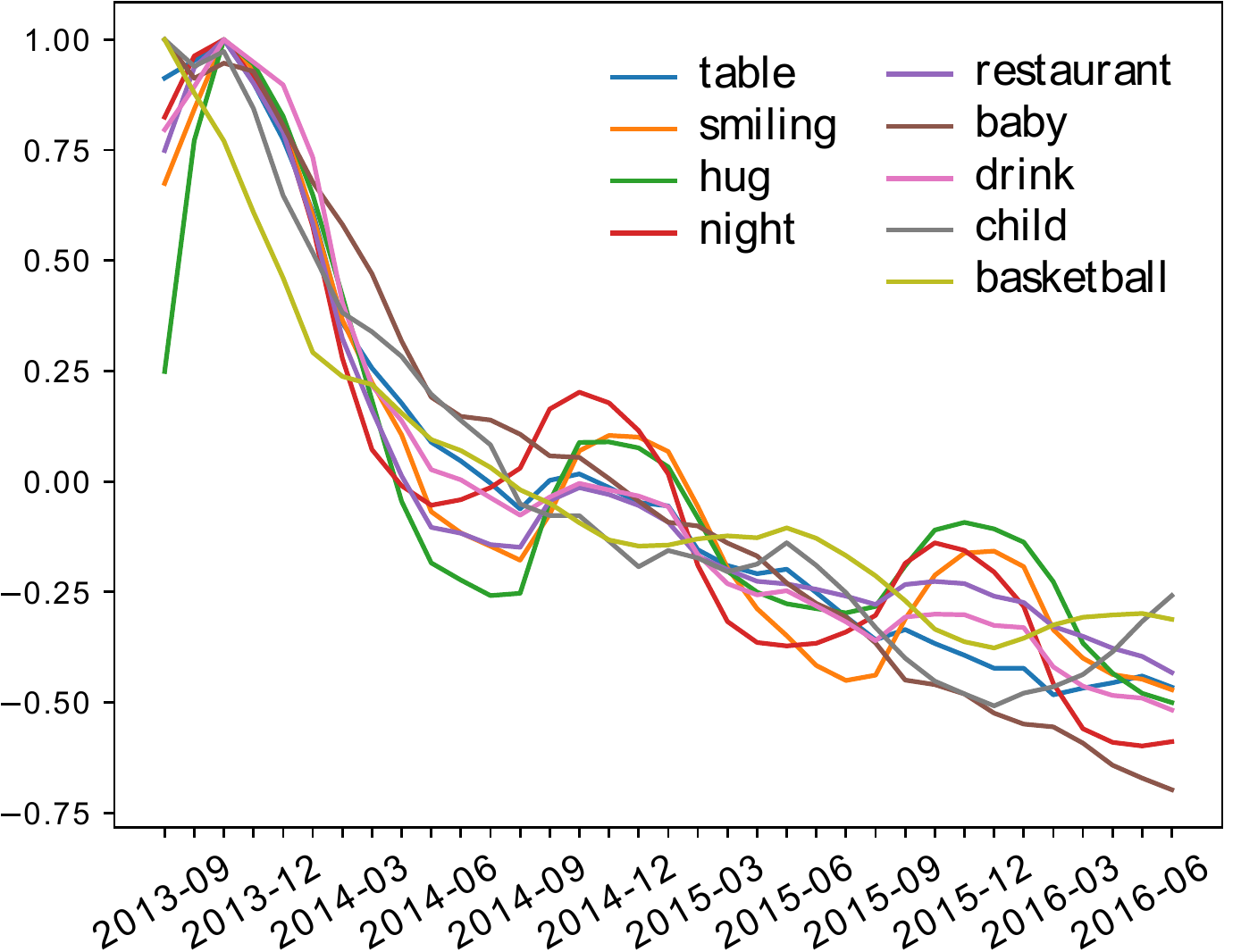}
 \caption{Decreasing}
 \end{subfigure}
 \begin{subfigure}[b]{.32\textwidth}
 \includegraphics[width=\textwidth]{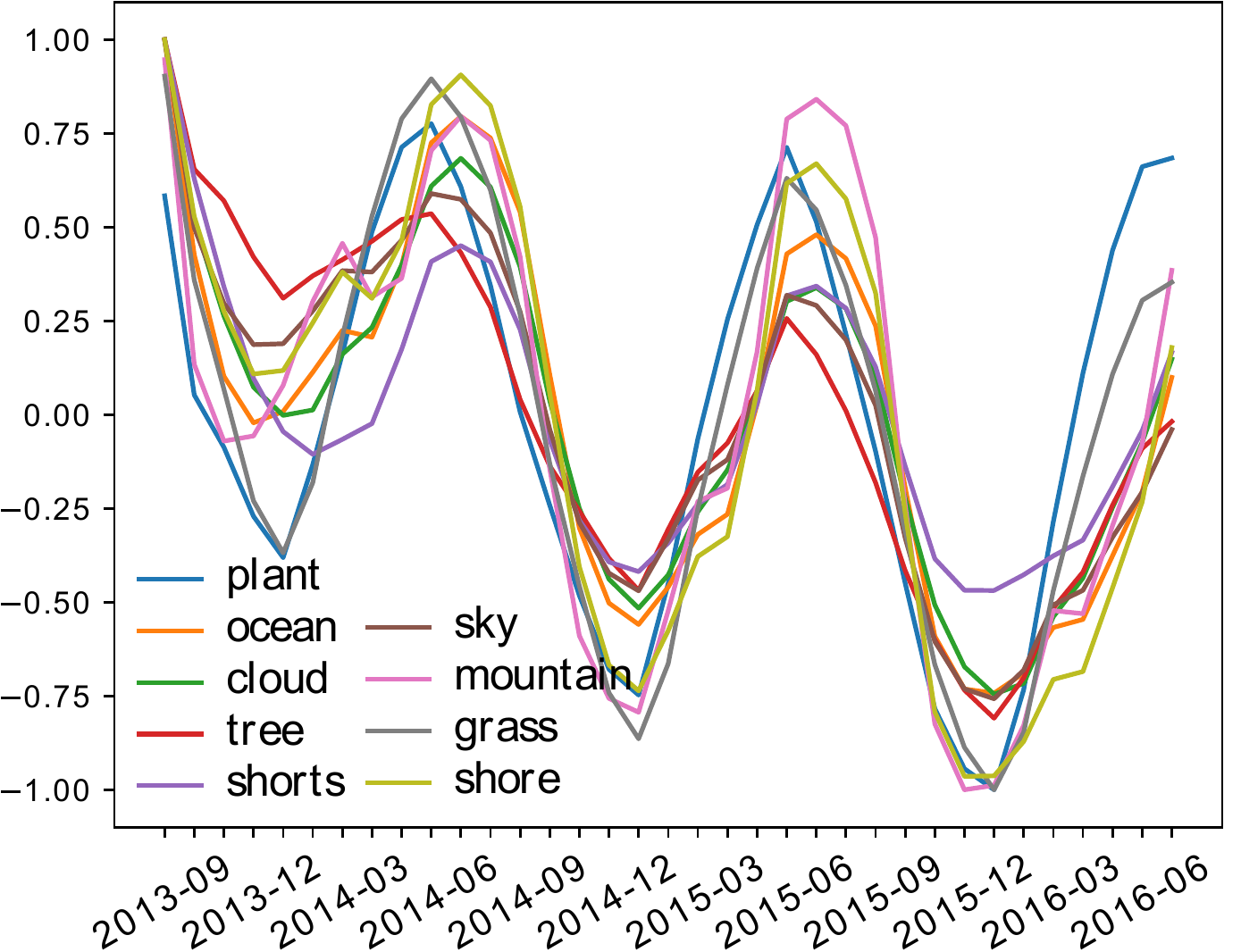}
 \caption{Seasonal}
 \end{subfigure}
\caption{
Three groups of visual concepts clustered by their temporal trends: (a) increasing, (b) decreasing, (c) seasonal variation. The y-axis represents the normalized concept popularity in the range $[-1, 1]$.
}
\label{fig:temporal}
\end{figure*}

\section{Classifying Photographs}
\label{sec:concepts}
\subsection{Visual Concepts}

\begin{figure*}[!htbp]
\centering
 \includegraphics[width=0.95\textwidth]{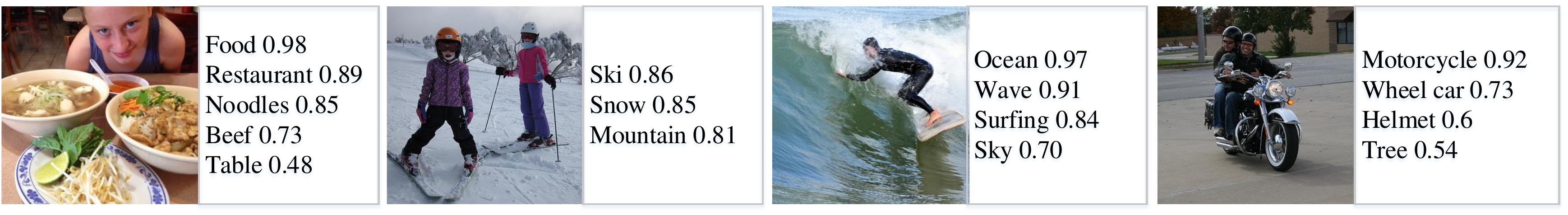}
\caption{Example images with top detected concepts and their scores. These images are not Facebook images but selected from a public dataset \cite{lin2014microsoft} for the purpose of displaying.
}
\label{fig:examples}
\end{figure*}

We are interested in recognizing many different types of cultural lifestyles or activities in photographs. To quantify such lifestyles we first need to identify the list of visual concepts that our classifier can learn to recognize. From indefinitely many candidate classes encompassing various human activities, we select the most common concepts ($K = 920$) organized in a 2-layer hierarchical structure including the following 11 categories in Table~\ref{tab:concept}. We provide the rationale and the full procedure to obtain the list as follows. 

What do we mean by culture? As stated earlier, we focus on common human activities in our daily lives. Therefore, we paid our attention to the common concepts portrayed in user photographs and took a bottom-up approach to construct the whole list of concepts. Specifically, we randomly sampled about 100k photographs and asked annotators to describe the main visible concepts of images using a few keywords. The obtained responses ranged from \textbf{objects} (\eg, car or banana), \textbf{actions} or \textbf{activities} (\eg, climbing or jumping) to scene attributes or even famous \textbf{places} (\eg, Opera House). After pruning infrequent keywords, we manually examined the whole set of keywords to merge redundant or similar concepts. 

We excluded keywords which are not strongly tied to apparent visual features or subjective expressions, such as `happy' or `fantastic' and potentially sensitive concepts (ethnicity, etc). However, we did not remove every concept which may not look directly relevant to ``culture'' such as `table' or `grass.' This is because such trivial objects still may indicate \textbf{events} (\eg, `picnic'), \textbf{interests} of users (\eg, 'home decoration') or \textbf{style} (\eg, `selfie'); these are very important to capture. Given the final list of concepts, we manually group them into 11 semantic categories. 

\subsection{Model and Training}
We collected annotations to train our model by an iterative approach. Human annotators provided binary (yes/no) annotations for each concept given an image. We start by annotating relatively a small number of photographs and train an initial model. Then we apply the model back to random image samples to seek hard negatives and hard positives and retrain the model. This procedure is repeated until the model achieves a robust classification accuracy. The annotators were instructed to focus on main concepts and ignore concepts which are very small or not clearly visible. The trained model thus follows the same behavior. 

We pose our problem as multiple binary classification instead of multiclass classification (1-out-of-K) such that our classes do not compete with each other. This also means that an image may have more than one concepts detected. See Figure~\ref{fig:examples} for example outputs of our model. The images were selected from a public image dataset for the privacy issue; but they resemble common images in Facebook.

To classify visual concepts from images, we use a deep residual network (ResNet-50) \cite{he2015deep}, which has shown the state-of-the-art performance for image classification. We train our model from scratch and take an iterative active learning approach as stated above. In addition, we replace the last softmax layer of the residual net for final classification with Sigmoid functions to perform multilabel classification. We crop the center region of an image and scale it to the canonical size of 224 by 224 pixels as in the standard practice. Each image takes around 200 ms to process in a single CPU. Our implementation is based on Torch. We follow most details and hyper-parameters specified in the original paper; See \cite{he2015deep} for the full details (github.com/facebook/fb.resnet.torch).

\section{Result}

\subsection{Concept Prediction Accuracy}
Table~\ref{tab:perf} presents the performance of our train models measured by area-under-curve (AUC) in ROC curves. Due to the space limit, we only show the aggregated performance grouped by each category and the average ratio of positive examples. The performance was measured on a completely separate set of images with more than 7M annotations, which were not used in training. 

\begin{table}[!htbp]
\centering{
\begin{tabular}{|l|p{1.4cm}|l|p{2.0cm}|}
\hline
Category & \# of Concepts & Avg AUC & Avg ratio of positives \\ \hline
Sports	&	64	&	0.972	& $1.93 \times 10^{-4}$ \\ \hline
Animals	&	108	&	0.982	& $3.14 \times 10^{-4}$ \\ \hline
Clothes	&	88	&	0.882	& $1.71 \times 10^{-3}$ \\ \hline
Food	&	107	&	0.979	&	$2.56\times 10^{-4}$ \\ \hline
Furniture	&	38	&	0.942	&	$5.52\times 10^{-4}$ \\ \hline
Music	&	16	&	0.983	&	$1.30\times 10^{-4}$ \\ \hline
Plants	&	33	&	0.954	&	$2.36\times 10^{-3}$ \\ \hline
Structures	&	17	&	0.973	&	$8.33\times 10^{-4}$ \\ \hline
Places	&	73	&	1.000	&	$6.03\times 10^{-6}$ \\ \hline
Scenes	&	113	&	0.923	&	$1.41\times 10^{-3}$ \\ \hline
Vehicles	&	53	&	0.978	&	$6.55\times 10^{-4}$ \\ \hline
\end{tabular}
}
\caption{Accuracy of visual concept classification.}
\label{tab:perf}
\end{table}

\subsection{Spatio-Temporal Trends}
\label{sec:sttrend}
Figure~\ref{fig:global} shows the global popularities of various concepts measured from photographs posted from 2013 July and 2016 June. For each concept, we obtained an average score per country while ensuring each country has at least 100,000 images per year during this period. As seen in this figure, some concepts (e.g., basketball) are ubiquitous and gaining a global popularity while some other concepts (e.g., American football) are concentrated on specific regions. 

As expected, many concepts reflect their actual spatial popularities (\eg, American Football or noodle). While we do not have ground truth to verify the accuracy, we observe the result exhibits similar patterns with a public index \cite{hecht2012explanatory}, which estimates the spatial relevance of concepts from Wikipedia data. However, not all concepts are strongly related to their origins or actual usage. For instance, the concept of `latte' shows a relatively small correlation with the actual coffee consumption per capita data ($r = .29, p < 0.001$)\footnote{www.caffeineinformer.com/caffeine-what-the-world-drinks} where East Asian countries tend to post the concept more frequently than Scandinavian countries, who in fact consume much more coffee. This suggests people may post photographs selectively according to their preferences or local trends.

\begin{figure}
\begin{tabular}{c}
\includegraphics[width=3in]{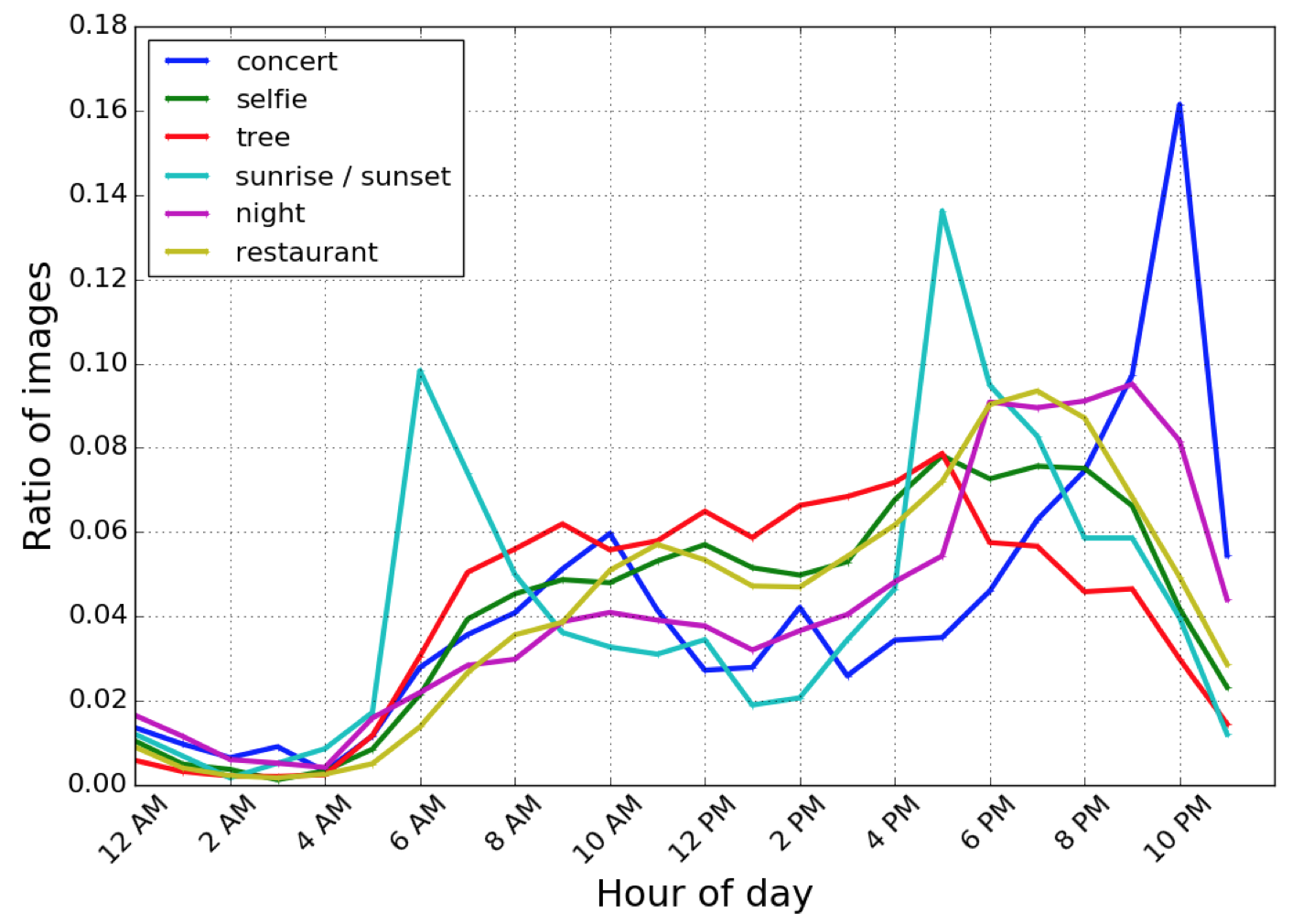} \\
\end{tabular}
\caption{
Temporal variations of visual concept popularities during a day in the UK.
}
\label{fig:ushour}
\end{figure}

We also examine the temporal changes or trends of the visual concepts. Figure~\ref{fig:temporal} shows three different patterns of trends: (a) increasing, (b) decreasing, (c) seasonal variation. We use dynamic time warping and K-means algorithm ($K = 7$ in this case; using a different $K$ did not significantly affect the obtained main patterns.) to cluster concepts based on their normalized temporal evolutions. Many seasonal concepts reach their peaks at a particular season, either summer or winter, and are suppressed in other seasons. The length of such a cycle might be annual for seasonal concepts such as ocean or skiing or daily for certain concepts such as night or restaurant (Figure \ref{fig:ushour}).

\begin{figure}
\centering
\begin{subfigure}[b]{.23\textwidth}
 \includegraphics[width=\textwidth]{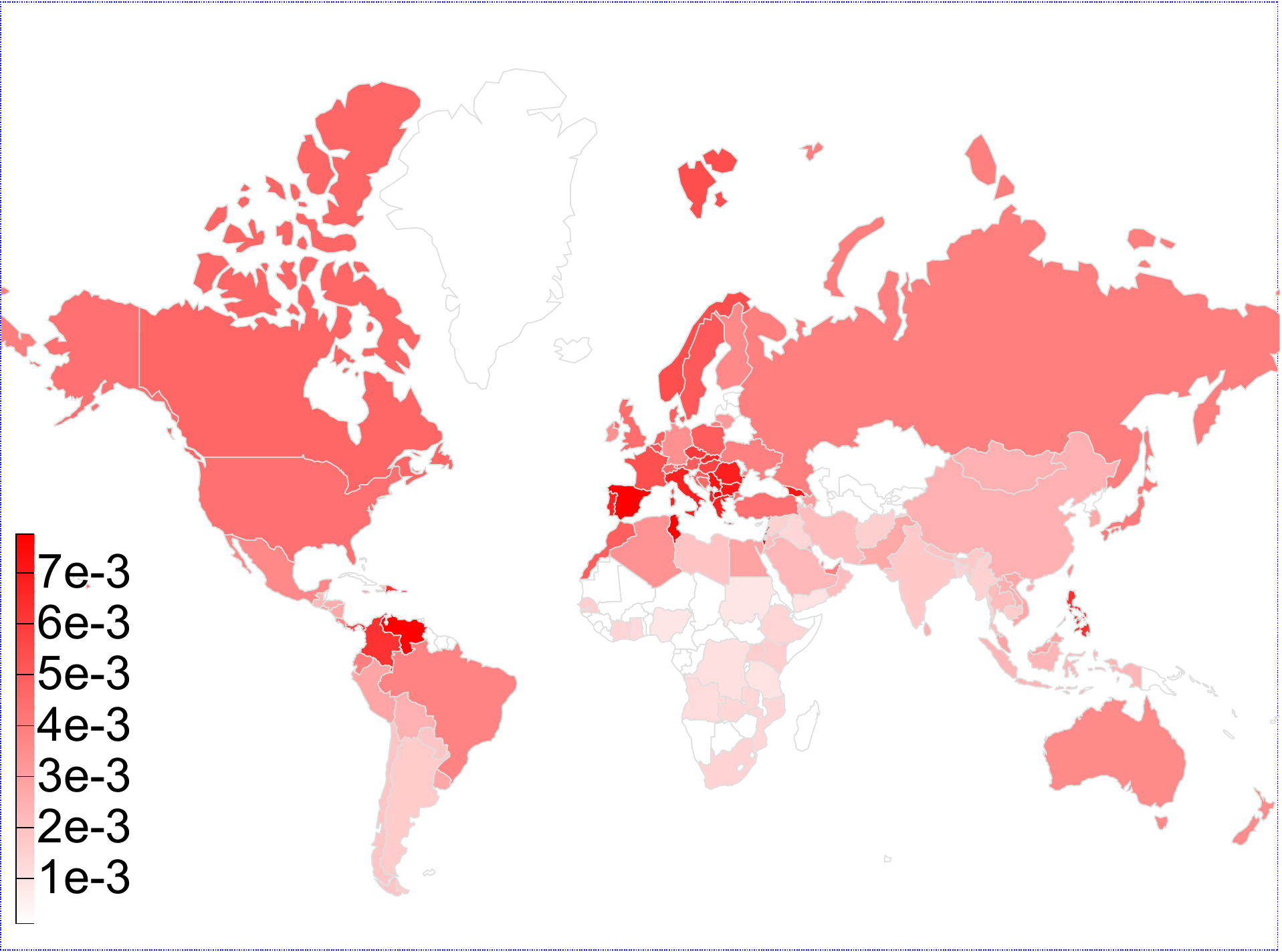}
 \caption{Swimming 2014 Aug}
 \end{subfigure}
 \begin{subfigure}[b]{.23\textwidth}
 \includegraphics[width=\textwidth]{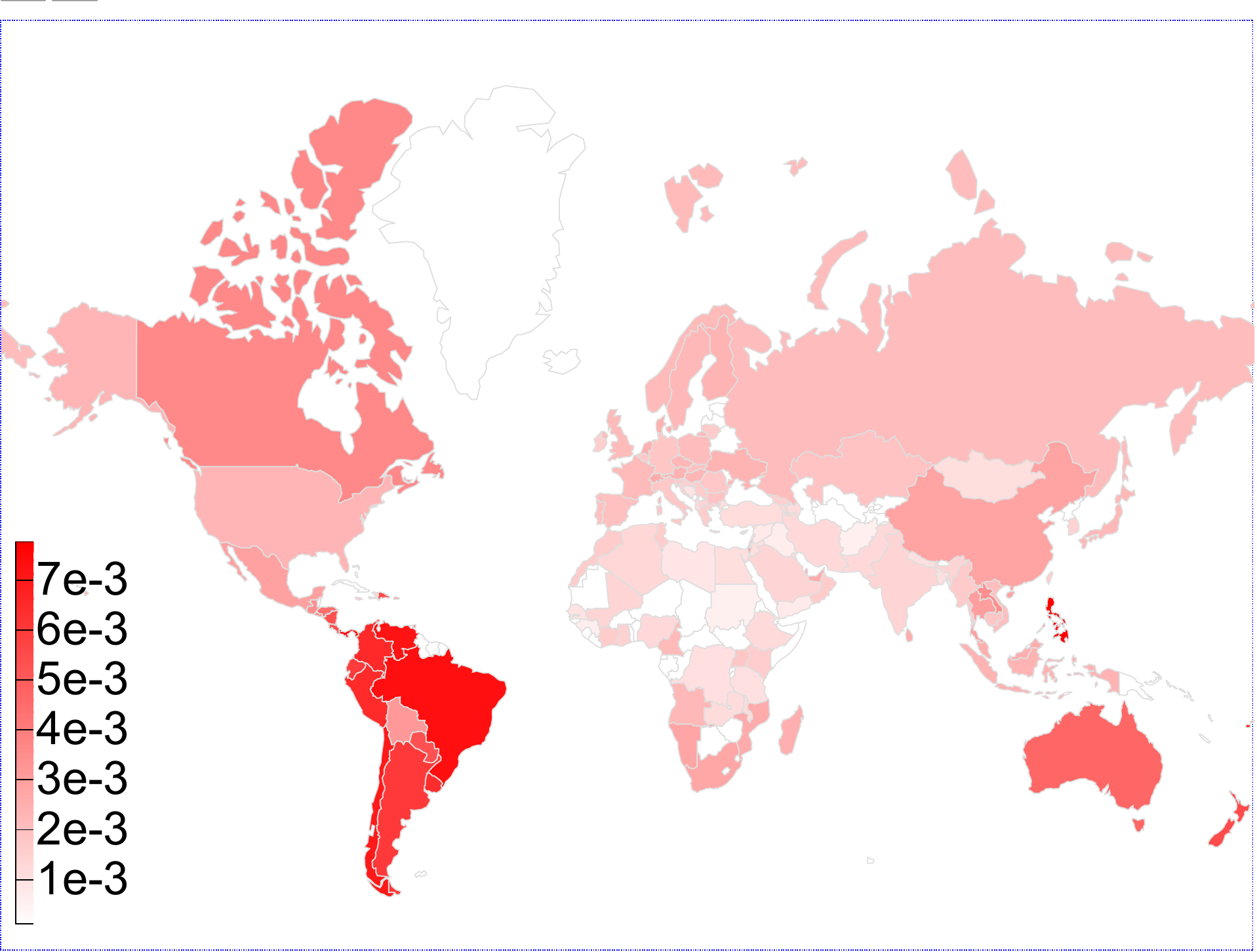}
 \caption{Swimming 2015 Feb}
 \end{subfigure}
 \begin{subfigure}[b]{.23\textwidth}
 \includegraphics[width=\textwidth]{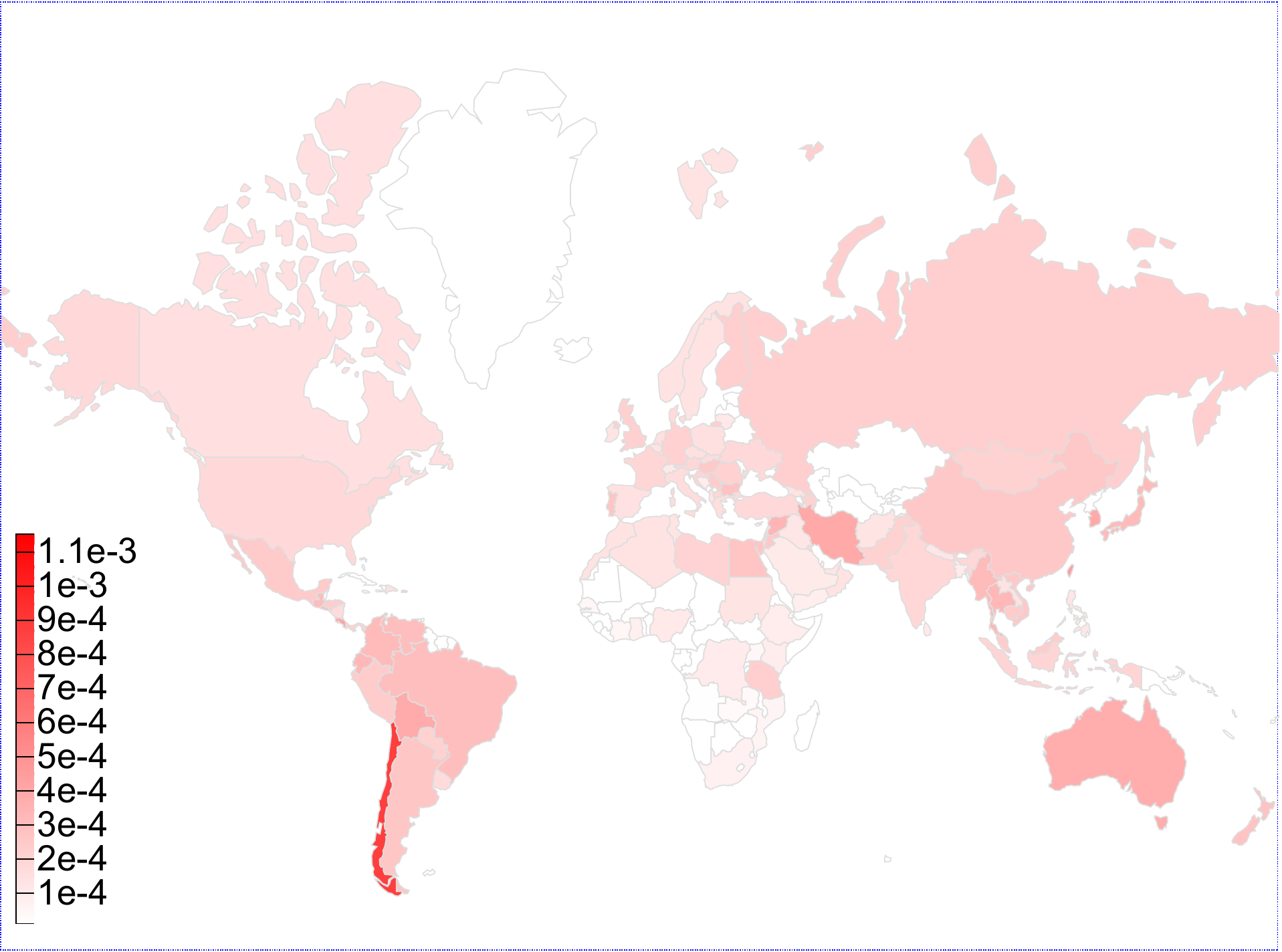}
 \caption{Snowman 2014 Aug}
 \end{subfigure}
 \begin{subfigure}[b]{.23\textwidth}
 \includegraphics[width=\textwidth]{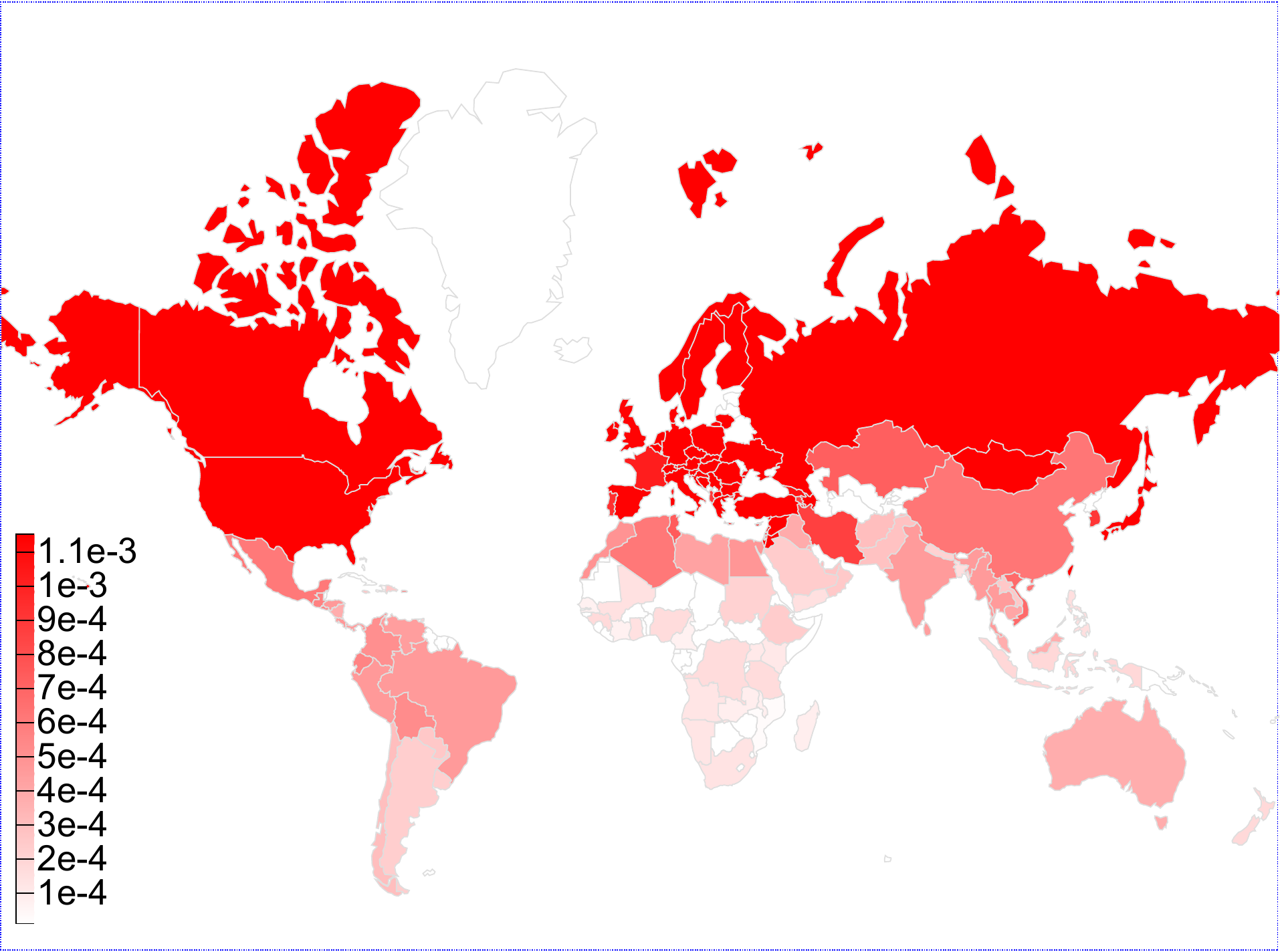}
 \caption{Snowman 2015 Feb}
 \end{subfigure}
\caption{
The temporal variations of seasonal concepts are often opposite in the Northern hemisphere and the Southern hemisphere.
}
\label{fig:spatemp}
\end{figure}

Lastly, we also present spatio-temporal co-evolutional visual trends. The popularities of some visual concepts change over time and location. The most common type of such patterns is again seasonal variation, where the Northern and the Southern hemisphere exhibit opposite patterns and alternate their status (\ie, active and inactive) on these concepts as shown in Figure~\ref{fig:spatemp}.

\subsection{Cultural Similarity between Countries}
From the presented visualizations, we note that the popularities of visual concepts differ from one country to another and it reflects local cultural or geographical factors. To investigate to which degree this distribution can characterize various local cultures of each country, we examine whether countries from similar cultural backgrounds (\eg, Western or Asian) also exhibit similar patterns of popular visual concepts among their user photographs.

\begin{figure}
\centering
\includegraphics[width=.43\textwidth]{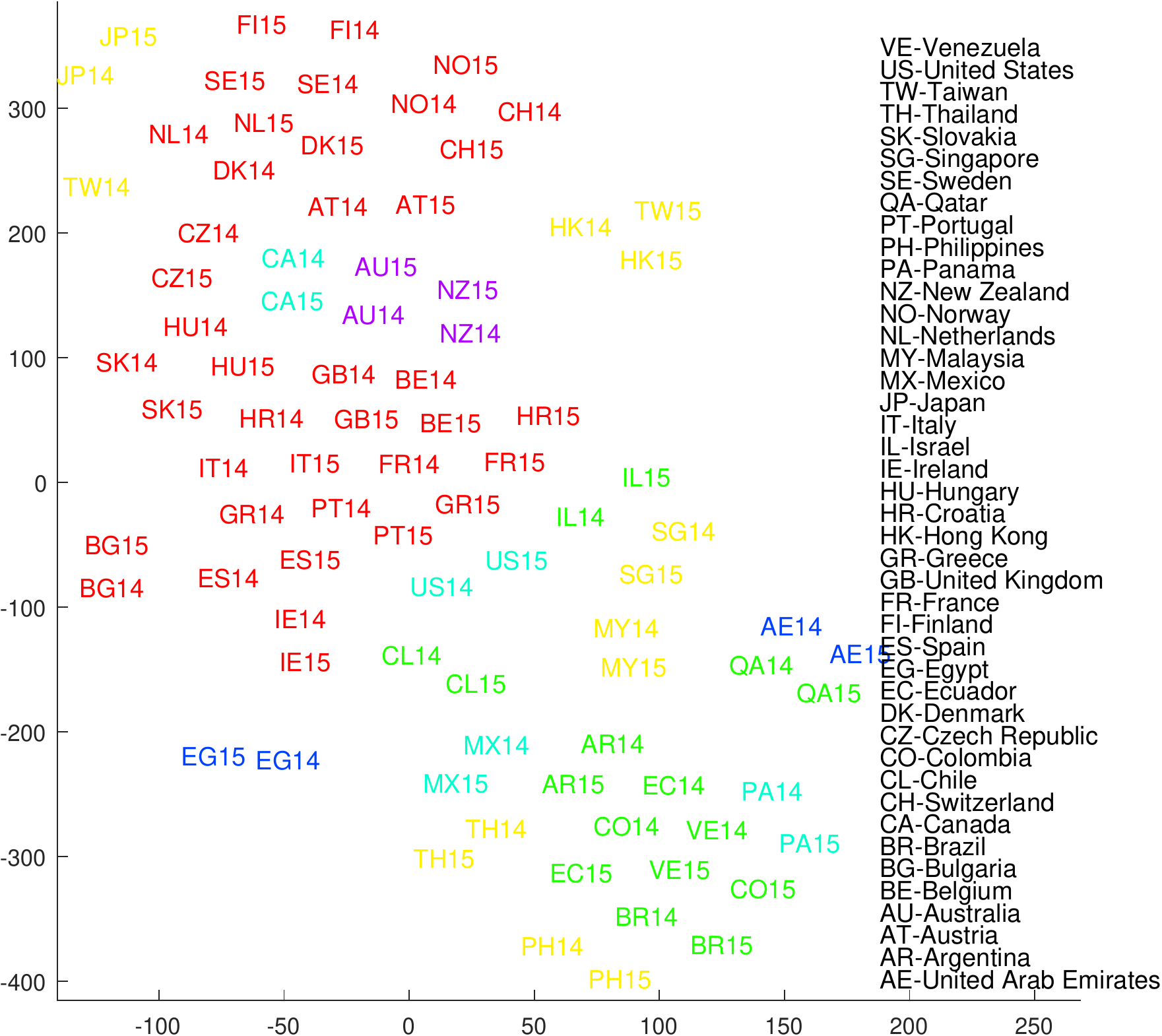}
\caption{
A t-SNE embedding of countries from inter-country visual similarities measured at 2014 and 2015. 
}
\label{fig:embedding}
\end{figure}

We first measure the similarity in cultural lifestyles in user photographs between countries. We estimate an average popularity of visual concepts for each country and use a cosine similarity between countries to obtain their visual similarities. To visually examine the inter-correlations between countries, we employ t-SNE~\cite{maaten2008visualizing} to map the countries into a 2-D plane while maximally preserving their inter-similarities as shown in Figure~\ref{fig:embedding}. We use the same color for the countries in the same continental region. We find that the countries from the same continent or from the similar cultural background (\eg, US and other European countries) are placed closely. The result indicates that the users' photographs convey the cultural lifestyles within each country. We also note that the embedding of the same country at different years (2014 and 2015) are very close, which suggests that the temporal variations tend to be smaller than inter-country variations.

In addition, we examine whether the visual similarity is correlated with socio-economic or geographical factors such as language, geolocation, GDP, etc. We obtain the social variables of the studied countries from an independent operated website (http://www.aneki.com/comparison.php). To simplify the estimation, we assign a binary attribute value to each social variable for a country pair to indicate whether two countries fall into the same category (\ie, speak the same language or in the same continent).
Then we measure individual Pearson's correlation coefficients between the social variables and the visual similarities, as shown in Table~\ref{tab:corr:pearson}.
All the cultural or socio-economic attributes considered are correlated with the visual similarity although none of single attributes yields a particularly strong correlation. This suggests that there might be multiple underlying cultural factors shaping what people commonly post in social media.

\begin{table}
\centering

\begin{tabular}{|l|l|l|}\hline
Social Index & $r$ & p-value \\ \hline
Climate	&	0.173	&	$<$ 0.00001 \\ \hline
HDI	&	0.215	&	$<$ 0.00001 	\\ \hline
GDP per capita	&	0.279	&	$<$ 0.00001 	\\ \hline
Languages	&	0.137	&	$<$ 0.00001 	\\ \hline
Religions	&	0.183	&	$<$ 0.00001	\\ \hline
Location	&	0.158	&	$<$ 0.00001	\\ \hline
\end{tabular}
\caption{Correlations between the visual similarity and socio-economic statues between countries (HDI: human development index).}
\label{tab:corr:pearson}
\end{table}

\subsection{Photograph similarity and friendship ties: Diffusion or Homophily?}
\begin{figure}
 \centering
    \includegraphics[width=.45\textwidth]{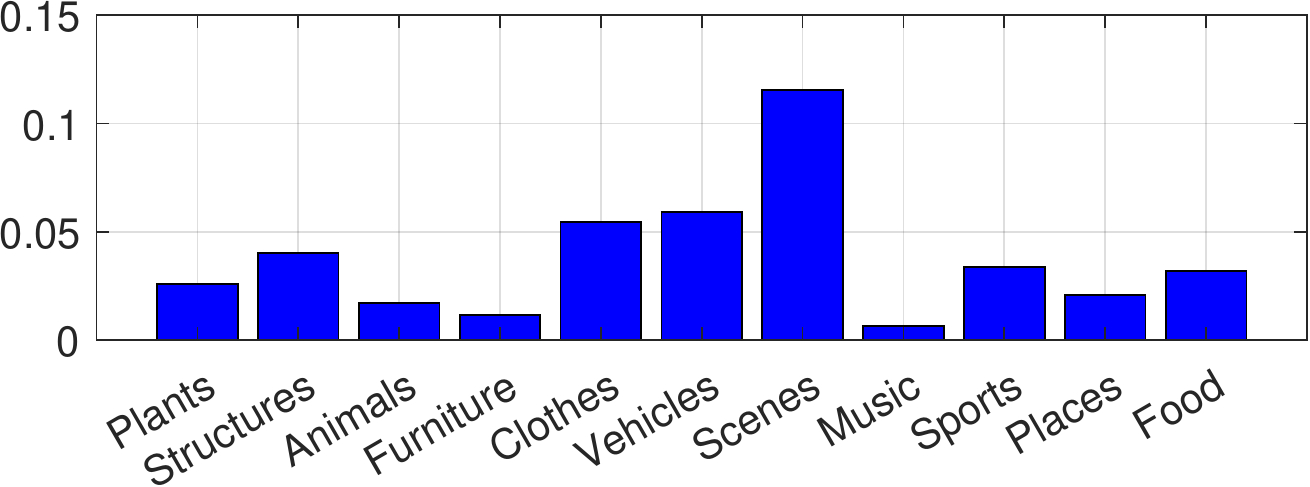}
 \caption{The difference of social correlation between friends and non-friends groups (gender and age controlled for). The mean of the friend group is significantly larger than the non-friend group except Music. All statistically significant (p-val $<$ 0.00001) except Music (p-val = 0.111).
}
 \label{fig:F_NF:corr}
\end{figure}
We now turn our focus to the second set of our research questions on the photograph similarity among Facebook users and its relation to the social ties, i.e., friendship. Prior research has suggested that tied individuals in a social network more likely share common behaviors or actions. Two popular explanations are (1) people with similar preferences might tend to become friends more easily (homophily); and (2) behaviors might be diffused through the social ties from one friend to another (influence). Diffusion of culture, innovations, and ideas has long been considered as a core function of mass media such as TV or movies. We wish to examine whether the propagation of culture can be also facilitated in online social media space.

To rigorously measure the causal effect of this procedure typically requires a manipulative experimental design with randomized interventions. We limit our scope in this paper to a purely observational study which does not build on any artificial controls over user activities or news feed. Therefore we use several statistical tests which can be used to infer suggestive effects of the network structure to social correlation, instead of a definitive causal inference.

We used our second data collection (Seattle) for the analysis in the section to rule out the confound of user location. For example, people who live close to each other would be more likely friends and also post more similar photographs due to local factors such as climate, local events, or any other local cultures. Although the granularity of city level might be considered coarse to rule out geographical confound completely, this was the finest scope to which we had access. We further control for user age and gender in the following analysis when applicable. The dataset comprises 1.3 M users and 250 M photographs posted by them from 2013 to 2016.

\textbf{Measuring similarity of users. }
We use different tests with slightly different ways of treating or counting visual concepts. Some prior studies considered individual discrete behaviors such as an adoption of a game~\cite{aral2009distinguishing} or tagging a specific keyword~\cite{anagnostopoulos2008influence}. In this case, an user's behavior is a binary variable and the correlation is measured based on whether friends have the \textbf{\textit{same}} behaviors or not. We follow this way in the Shuffle test while treating each individual concept separately. There exist other studies which measure how \textbf{\textit{similar}} their behaviors or the content they generate are~\cite{sharma2016distinguishing}. We use this way as well in the PME test where we measure the visual similarity using all concepts and the following subsection of social correlation. 

\subsubsection{Social Correlation among Friends}
We first examine whether individuals linked with social ties (i.e., friends) post more similar photographs than people without direct ties.

Let's denote by $V$ the whole set of 1 M users and by $E$ the set of edges (friendship ties) between them. For each friend pair $(v_i, v_j) \in E$,
we randomly select another user $v_k \in V$ such that $(v_i, v_k) \notin E$ (non-friend).
Let's denote the set of the tuples of selected users $(v_i, v_j, v_k)$ by $T$.
Note that selecting $v_k$ needs to be done carefully because there might be external factors such as user's gender which would affect both the likelihood of friendship and that of the visual similarity.
To control for such confounding factors, we enforce $v_k$ to be of the same gender and age group as $v_j$.

Let $x_i$ be the vector of average scores of the visual concepts in each category of the user $v_i$. Then we measure cosine similarities of $(x_i, x_j)$ and $(x_i, x_k)$. Finally the social correlation here is defined by the average difference between these two similarities such that
\[
D_{\text{corr}}(T) =\frac{1}{|T|} \sum_{(v_i, v_j, v_k) \in T}{ \text{cos}(x_i, x_j) - \text{cos}(x_i, x_k). }
\]

\begin{figure}
 \centering
 \begin{subfigure}[b]{.23\textwidth}
 \includegraphics[width=\textwidth]{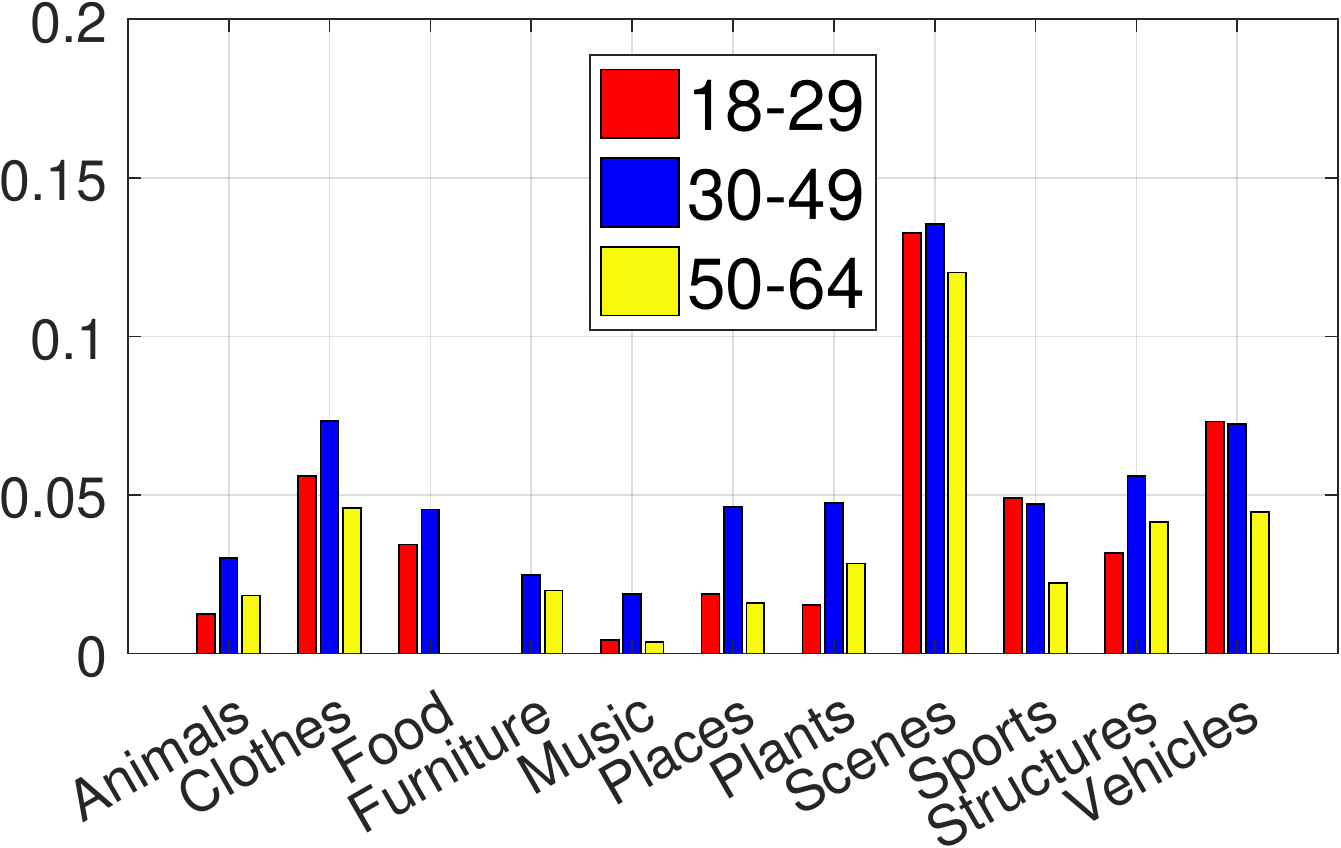}
 \caption{Different Age Groups}
 \label{fig:F:corr:age}
 \end{subfigure}
 \begin{subfigure}[b]{.23\textwidth}
 \includegraphics[width=\textwidth]{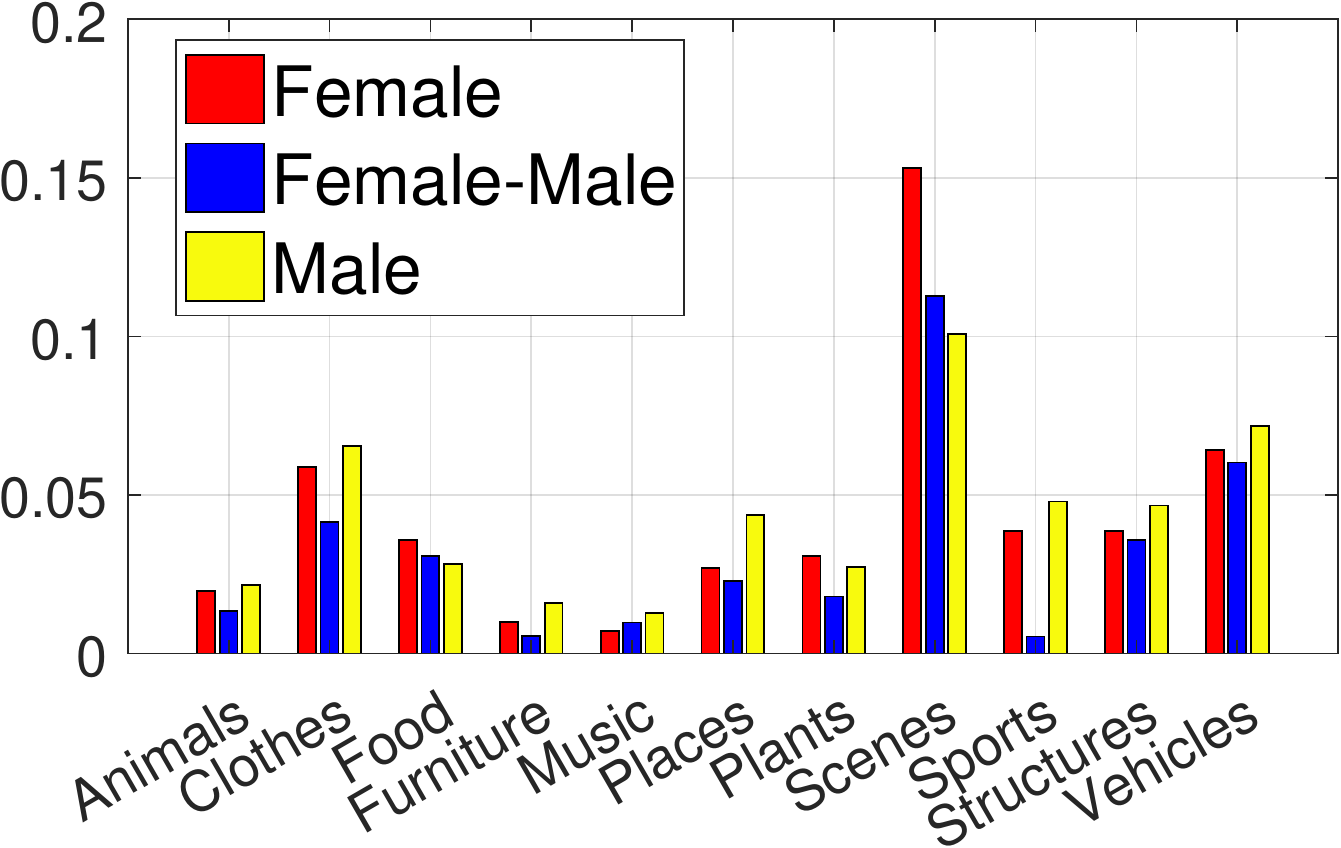}
 \caption{Different Gender Groups}
 \label{fig:F:corr:gender}
 \end{subfigure}
 \caption{The difference in social correlation between friends and non-friends pairs. The results are separated by user age and age group. }
 \label{fig:F:corr}
\end{figure}

\begin{figure}
\begin{tabular}{cc}
\includegraphics[width=1.6in]{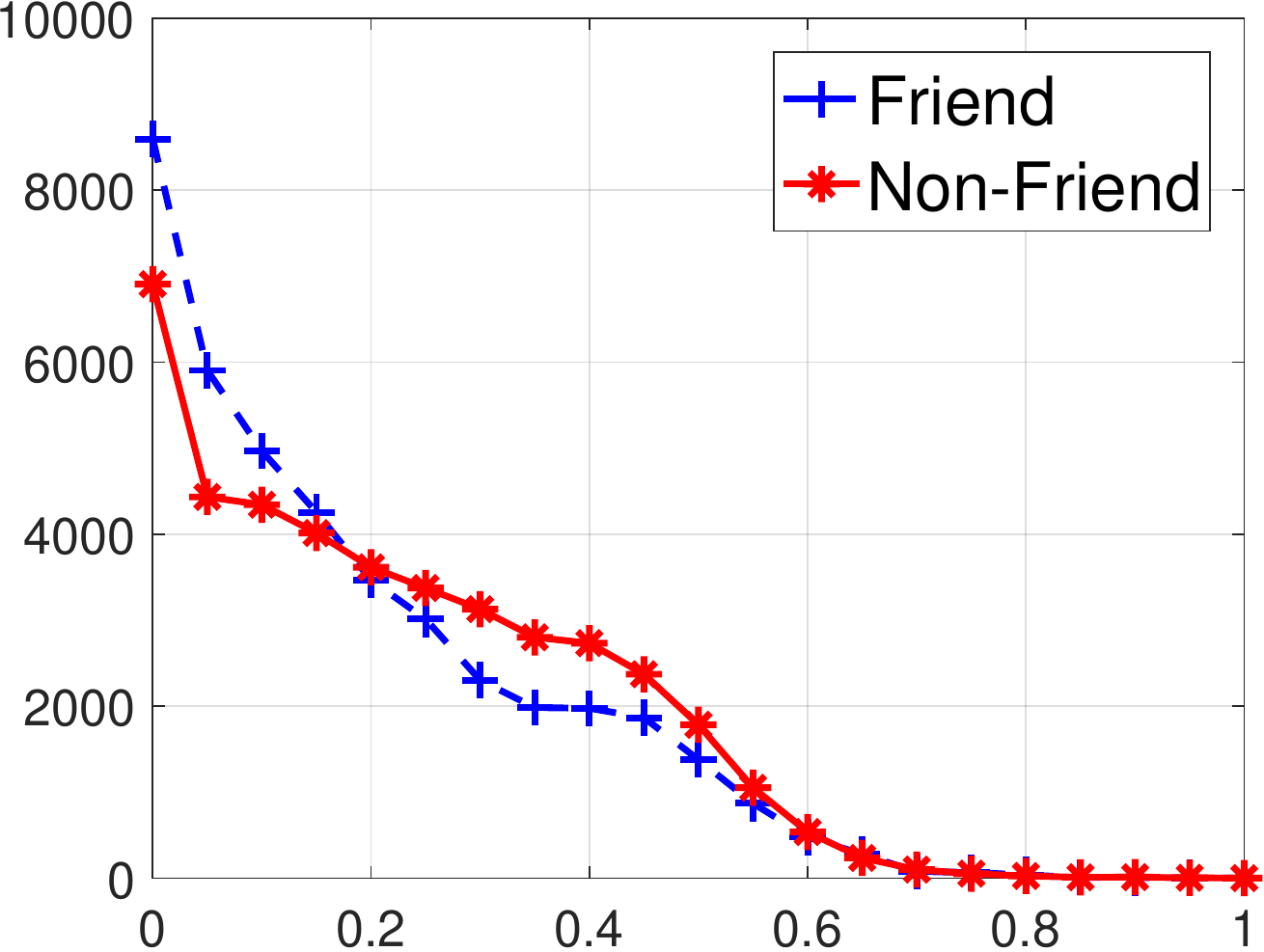} &
\includegraphics[width=1.6in]{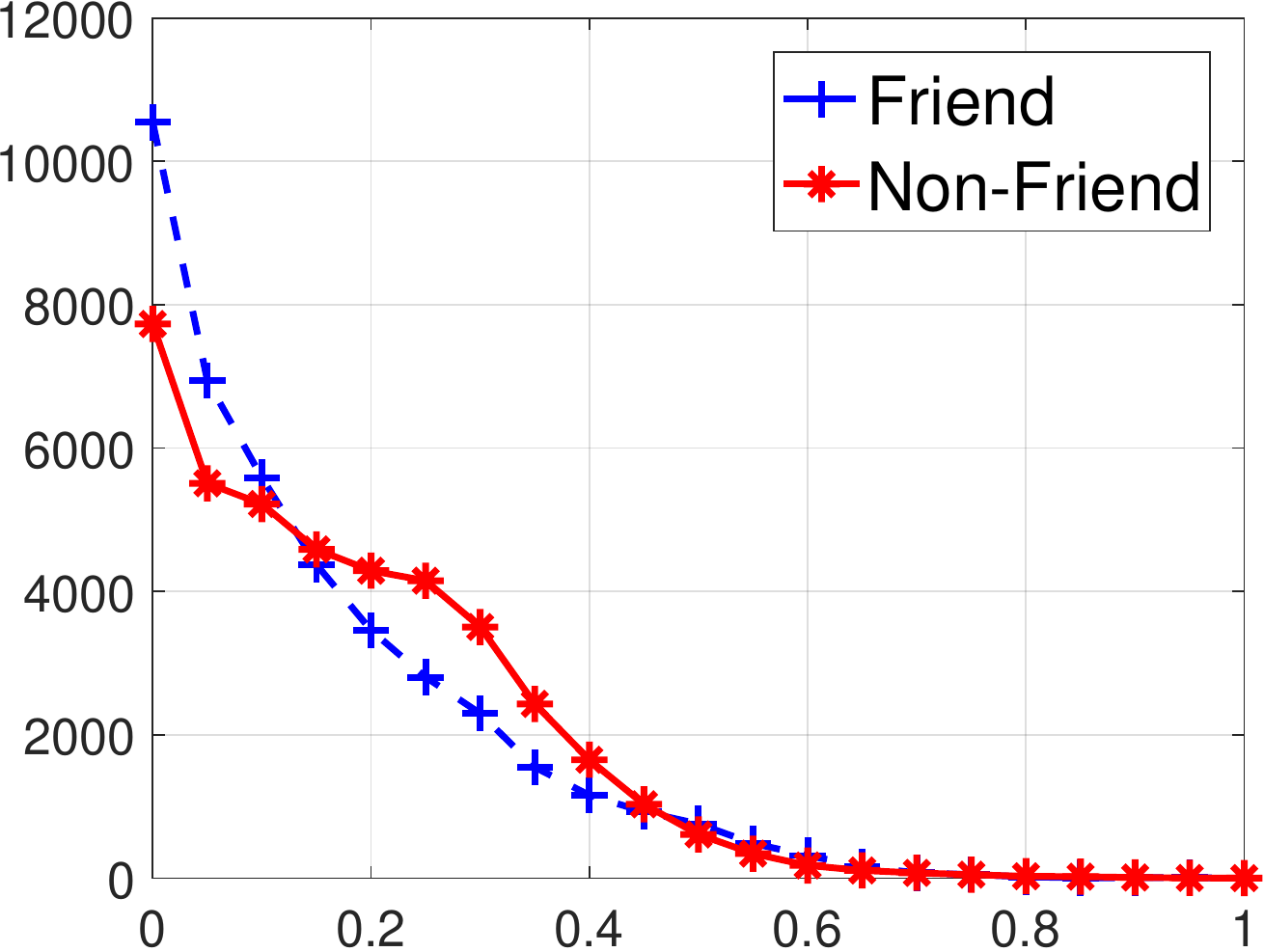} \\
(a) Selfie & (b) Smile \\[6pt]
\includegraphics[width=1.6in]{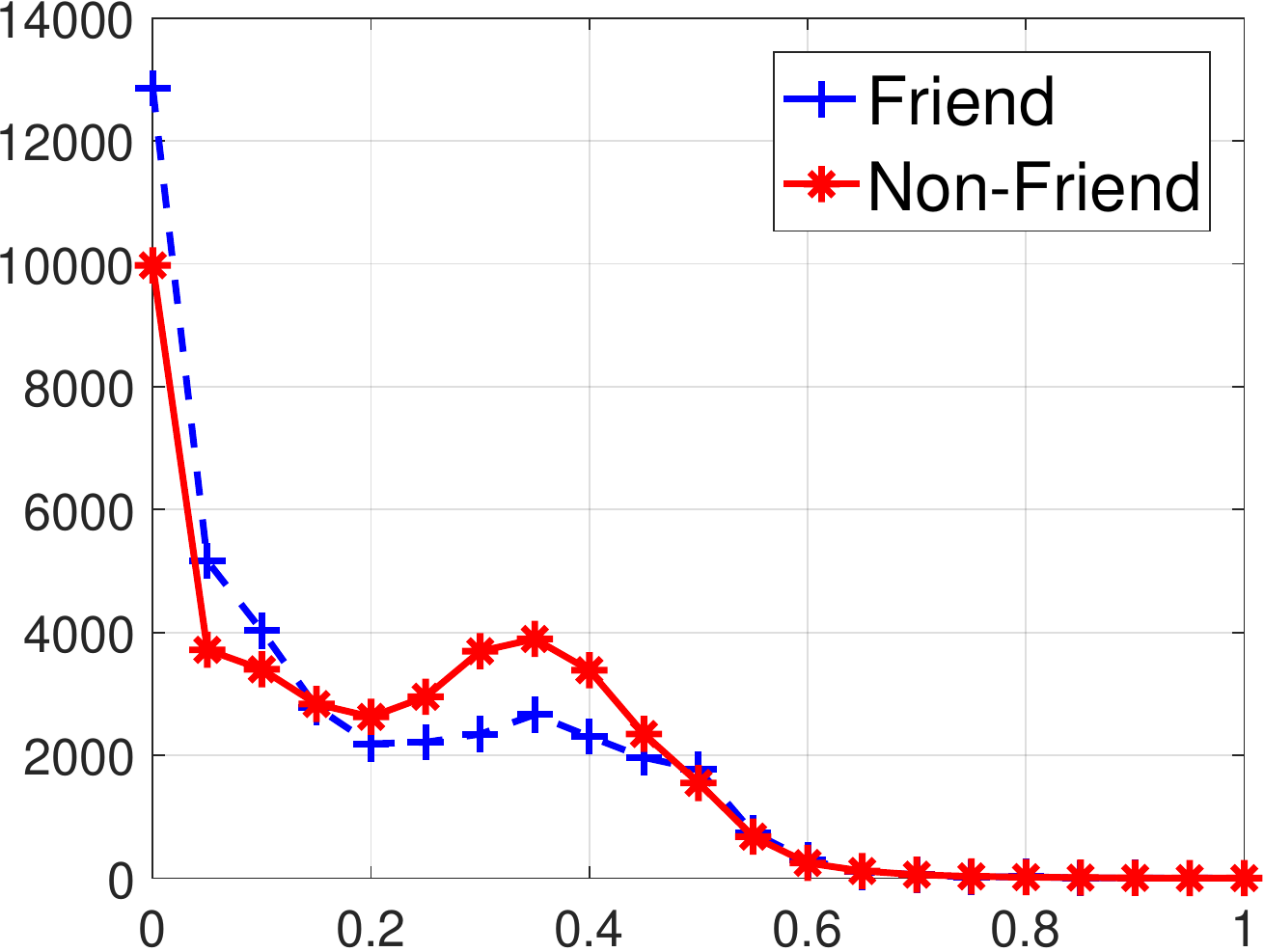} &
\includegraphics[width=1.6in]{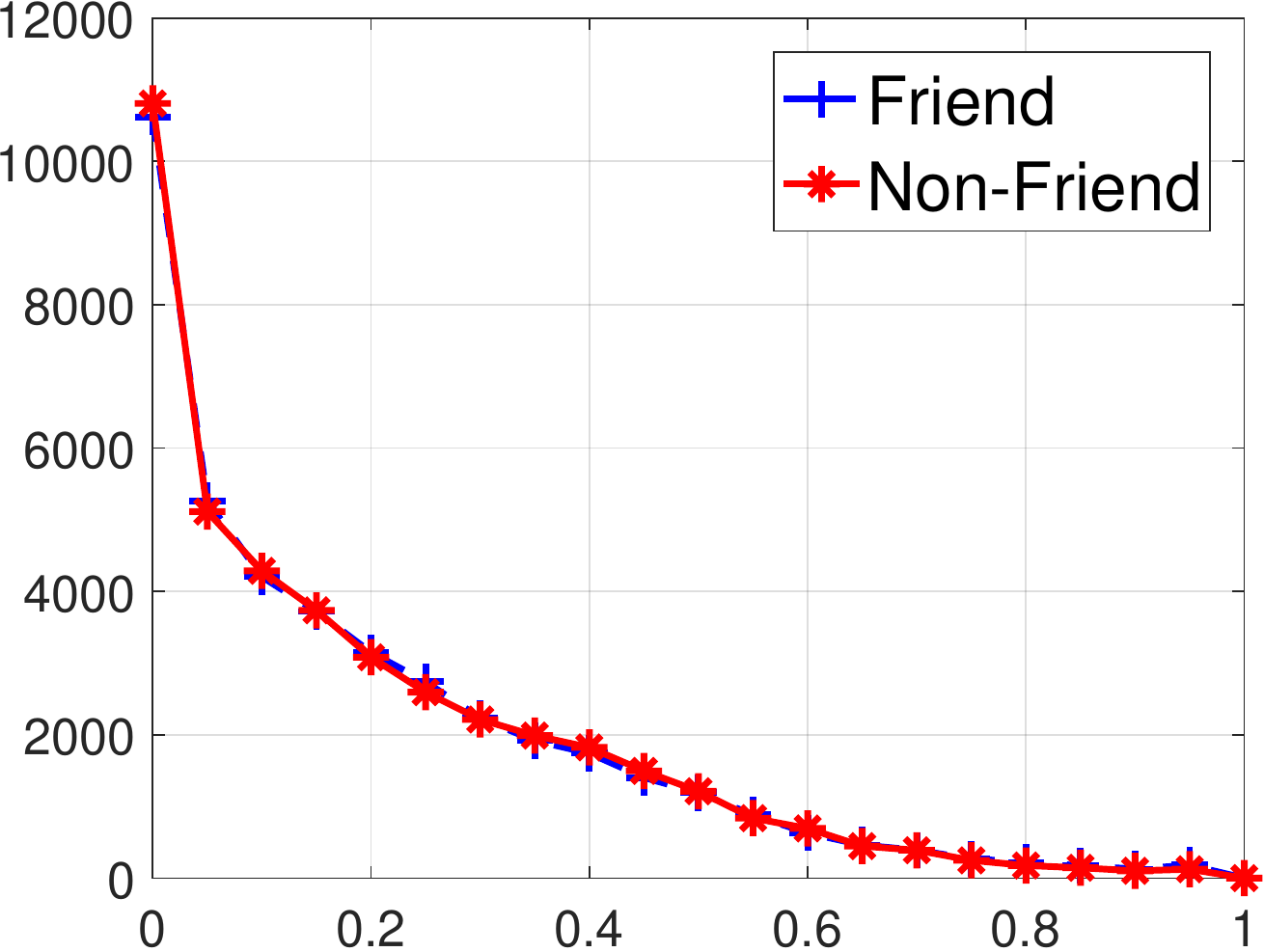} \\
(c) Jeans & (b) Sunglasses \\[6pt]
\end{tabular}
\caption{
Social correlation: histograms of the difference of the average scores between friend and non-friend pairs for 4 visual concepts. The x-axis represents the score difference and the y-axis represents the number of pairs. The friend pairs are more similar in selfie, smile, and jeans, and two groups are similar in sunglasses (no correlation).
}
\label{fig:socialcorr}
\end{figure}

Figure~\ref{fig:F_NF:corr} reports the difference of the social correlations between friend pairs and non-friend pairs across the 11 categories (Sec. Visual Concepts). 
We performed a t-test to see whether two groups are statistically distinct and found that the friend group has significantly larger correlations than the non-friend group across all categories except music. Figure~\ref{fig:socialcorr} shows the differences of the average concept scores between friend and non-friend pairs for 4 different concepts.

We also investigate whether the degree of social correlation differs by demographic groups. Figure~\ref{fig:F:corr} summarizes the differences in social correlation between friend pairs and non-friend pairs, reported separately for each age group or gender. We note that the users in 30-49 age group show bigger differences in correlations in most categories while younger users (18-29) are more correlated with their friends in vehicles and clothes. On the other hand, friends of the same gender are more correlated than non-friends of the same gender as well as friends of different genders, especially in Clothes and Sports.

\subsubsection{Predictive Diffusion: Shuffle Test}
We now apply statistical tests proposed in recent papers to verify the effects of social influence in driving the observed social correlations. The Shuffle test \cite{anagnostopoulos2008influence} is such a method to distinguish the source of the observed correlation. These methods compare social correlations between what is actually observed and what would have been observed if there had been no effects of social influence. These tests are not designed to infer a causal relationship and the term ``influence'' should be interpreted as \textit{predictive influence.}

The procedure starts by fitting a logistic regression model with the original data and estimating a correlation parameter for each concept. Then the timestamps of user actions (\ie, photograph post times) are randomly permuted. We then estimate the model parameters on the permuted data and compare them with the original parameters.

\begin{figure}
\begin{tabular}{cc}
\includegraphics[height=1.6in]{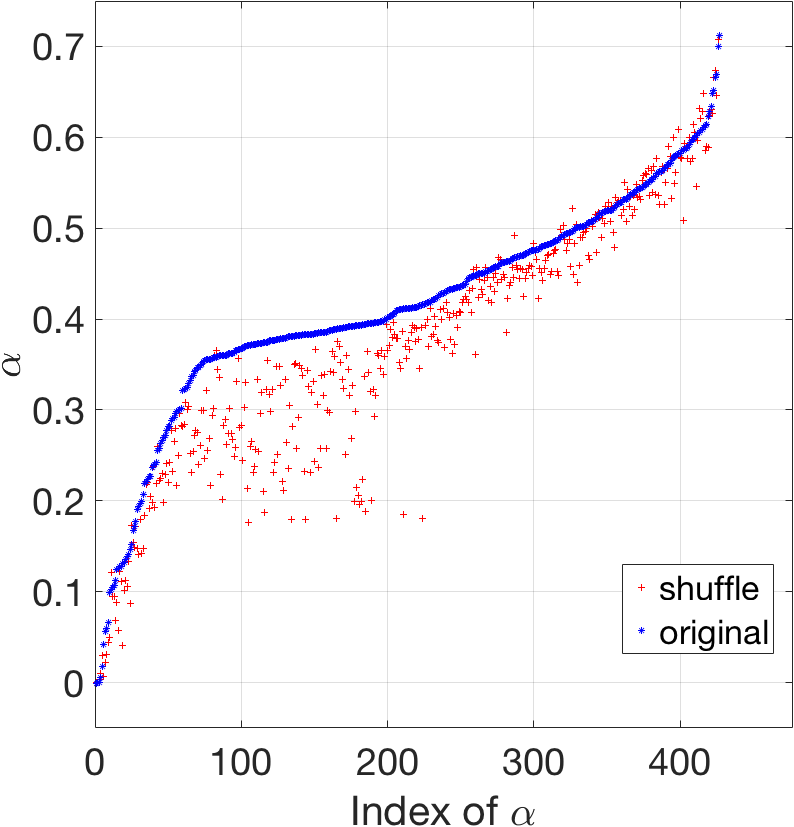} &
\includegraphics[height=1.6in]{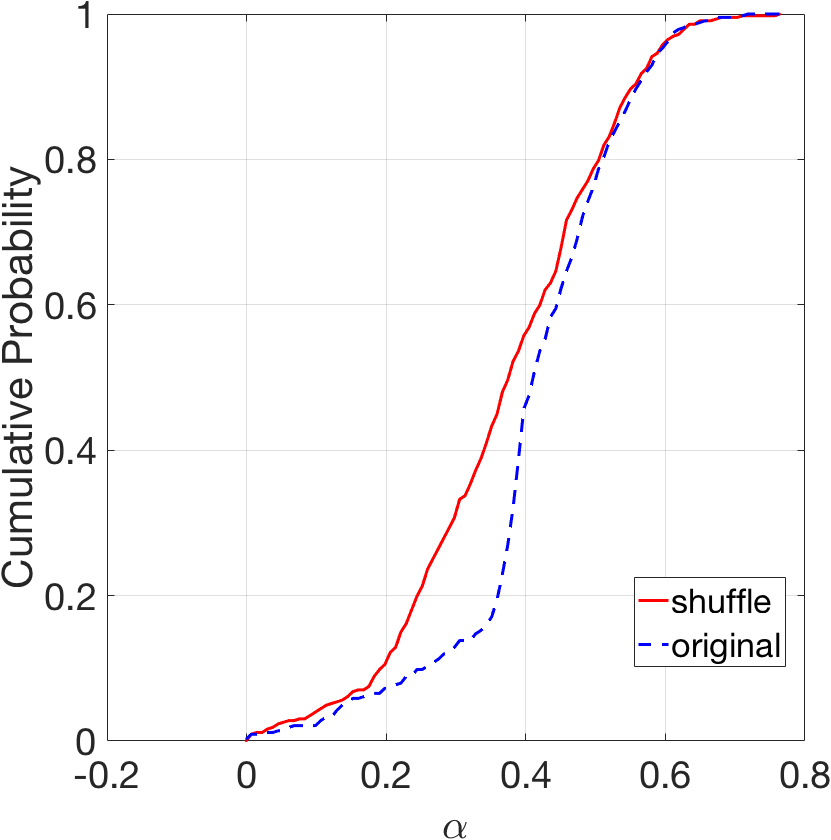}\\
(a) & (b)  \\[6pt]
\end{tabular}
\caption{
(a) A scatter plot showing the correlation coefficients estimated from original data (blue) and randomly permuted data (red).
(b) Cumulative density functions of the coefficients. These show the correlation is stronger in the original data.
}
\label{fig:shuffle}
\end{figure}

Figure~\ref{fig:shuffle} shows the distributions of the correlation coefficient, $\alpha$. A higher $\alpha$ means that the correlation is stronger and friends post more similar photographs. We can see this correlation is stronger in the original data before permutation. The mean values of all $\alpha$ before the shuffle was 0.413 (SD = 0.123) and after the shuffle was 0.371 (SD = 0.143) with a t-test verifying the former is larger (t=4.538, p-val $<$ 0.00001). This means that the sequence of actions of posting the same visual concept (of friends) is aligned with their friendship links. In other words, the decision of each user to post the concept or not is correlated with the number of friends who \textbf{\textit{recently}} share the same concept. This might be due to (1) the increased likelihood for an exposure and/or (2) the social ``threshold'' required for one's adaptation \cite{granovetter1978threshold}. Therefore, this result can be suggestive of the influential role of social ties in the diffusion of such visual concepts as opposed to homophily.

\begin{table}
\centering
\begin{tabular}{|c|l|l|l|}
\hline
Concept	&	Shuffled	&	Original	&	Difference	\\ \hline
face	&	0.18	&	0.416	&	0.236	\\ \hline
person	&	0.185	&	0.411	&	0.226	\\ \hline
child	&	0.18	&	0.389	&	0.209	\\ \hline
smiling	&	0.179	&	0.383	&	0.204	\\ \hline
table	&	0.179	&	0.381	&	0.202	\\ \hline
tree	&	0.196	&	0.393	&	0.197	\\ \hline
night	&	0.176	&	0.371	&	0.195	\\ \hline
sky	&	0.2	&	0.395	&	0.195	\\ \hline
pants	&	0.2	&	0.393	&	0.193	\\ \hline
hug	&	0.187	&	0.374	&	0.187	\\ \hline
shoes	&	0.206	&	0.393	&	0.187	\\ \hline
plant	&	0.214	&	0.392	&	0.178	\\ \hline
drink	&	0.21	&	0.374	&	0.164	\\ \hline
restaurant	&	0.202	&	0.36	&	0.158	\\ \hline
hat	&	0.221	&	0.379	&	0.158	\\ \hline
\end{tabular}
\caption{Top concepts with the largest $\alpha$ changes in Shuffle test. }
\label{tab:shuffle:top:diff}
\end{table}

Table~\ref{tab:shuffle:top:diff} further reveals the top concepts that have the largest correlation changes, before and after the shuffle. These concepts are more sensitive to the timestamps of users' behaviors, which suggests the correlation between friend users are more likely due to influence by their friends on these visual concepts. The concepts of `face' or `person' are not directly pertinent to culture; however, they can be highly indicative of other activities such as sports or group events. 
Also, Table~\ref{tab:shuffle:top} lists the concepts that have largest correlation coefficients, but are less sensitive to the timestamps of users' behaviors. Therefore, these high correlations in these concepts are likely mainly driven by homophily. This result also suggests that a high correlation does not always mean a contagion.

\begin{table}
\centering
\begin{tabular}{|c|l|l||c|l|l|}
\hline
Concept	&	S	&	O	&	Concept	&	S	&	O	\\ \hline
pumpkin	&	0.76	&	0.71	&	panties	&	0.59	&	0.62	\\ \hline
cosmetics	&	0.71	&	0.70	&	crying	&	0.59	&	0.61\\ \hline
truck	&	0.65	&	0.67	&	coffee	&	0.65	&	0.61		\\ \hline
watch	&	0.67	&	0.67	&	bread	&	0.63	&	0.61\\ \hline
handbag	&	0.67	&	0.65	&	juice	&	0.62	&	0.61		\\ \hline
meme	&	0.63	&	0.63	&	tv	&	0.63	&	0.61\\ \hline
\end{tabular}
\caption{Top concepts with the largest $\alpha$ before Shuffle test}
\label{tab:shuffle:top}
\end{table}

\subsubsection{Preference-based Matched Estimation Test}
We also use the Preference-based Matched Estimation (PME) test to distinguish the effects of influence and homophily \cite{sharma2016distinguishing}. Unlike the Shuffle test which operates on each individual visual concept, we now measure the overall similarity across all concepts. The PME was originally used to analyze the \textit{copy} of a particular behavior. Thus, the temporal order matters in this type of analysis. However, in our experiments, we are interested in the implicit propagation of behaviors -- the behavior of uploading a photo with similar content, and there are usually delays more than several days in similar posts. Therefore, our results are less sensitive to the perturbed temporal order introduced by Facebook's feed ranking algorithm.

The PME test assumes that there is no social influence between non-friend users and estimates the effect of influence by replacing a friend of a user with a non-friend similar to the friend.
Specifically, the PME test first tries to match each friend $f$ of user $u$ with a non-friend user $s$, who has similar preference with user $f$.
In the next stage, it estimates the social influence by subtracting the correlation between non-friend pair $s$ and $u$ from the correlation between friend pair $f$ and $u$.
We use the following two criteria to match user $f$ and $s$ at time $t$.
\begin{itemize}
\item{User $f$ and $s$ should post images having similar set of concepts. Following~\cite{sharma2016distinguishing}, we use Jaccard Index to compute the overlap between them: \[J(f,s) = \frac{|A_t^f \bigcap A_t^s|}{|A_t^f \bigcup A_t^s|},\] where $A_t^i$ is the set of concepts of the images posted by user $i$ before time $t$.}
\item{The number of concepts of f and the number of concepts of s should also be similar. We define it as \[N(f,s) = \frac{\left||A_t^f|-| A_t^s|\right|}{|A_t^f |}.\]}
\end{itemize}

We choose $J(f,s) > 0.9$ and $N(f,s) < 0.1$ in our implementation. Next, the influence is estimated as the difference of the correlation between $u$ and his friend $f$, and the correlation between $u$ and his non-friend $s$:
\[
\mathtt{Inf(u)} = \frac{\sum_{c\in A^{F(u)}_{t+1}}I(c\in A^u_{t+1})}{| A_{t+1}^{F(u)}|} -  \frac{\sum_{c\in A_{t+1}^{S(u)}}I(c\in A^u_{t+1})}{| A^{S(u)}_{t+1}|},
\]
where $F(u)$ is set of friend users of $u$ and $S(u)$ is the matched non-friend users, $ A^{F(u)}_{t+1}$ and $A^{S(u)}_{t+1}$ are the set of concepts in the images posted by the group of users $F(u)$ and $S(u)$ from time $t+1$ respectively, and $A^u_{t+1}$ is the set of concepts for user $u$ from time $t+1$.
We choose a timestamp $t$ in 2015/05, a year earlier from the latest time in our dataset. Next, we compute the concept set $A^u_t$ of images posted by user $u$ in the past half year before 2015/05 and the concept set $A^u_{t+1}$ as the past half year before 2016/05.

\begin{table}
\centering
\begin{tabular}{|l|l|l|l|}
\hline
Influence&	std &	t-stat&	p-value \\ \hline
0.07&	0.09&	293&	$< 10 ^ {-5}$ \\ \hline
\end{tabular}
\caption{Results of the PME test}
\label{tab:pme}
\end{table}
Table~\ref{tab:pme} shows the result of the PME test. Overall, we found that there exists an effect of social influence among user photographs. Although the variation across users tends to be large, the t-test verifies its statistical significance. This result also confirms that the effect can be found when considering \textit{all} concepts together. 

\section{Discussion and Conclusion}
We have shown that automated classification of visual content of photographs in social media is an effective means to assess the local and global trends of various cultural activities and lifestyles, depicted in user photographs in Facebook. We have built a scalable computational pipeline to process Facebook photographs and analyze various spatio-temporal patterns as well as its diffusion pattern via social ties.
We specifically focus on understanding how people communicate and interact \textit{visually} and did not use common features such as user texts or hashtags. The visual cues do not require any translation between countries speaking different languages, thus our analysis seamlessly applies across many different regions and cultures universally. Overall, our analysis suggests two important findings as follows.

Firstly, user photographs in Facebook display a variety of cultural lifestyles and preferences on many categories such as sports, food, fashion, etc. Inferring human activities from photographs is thus an effective way of understanding popular activities or preferences at specific places and times. The granularity of analysis is very fine as exemplified in our analysis during a day (Fig.~\ref{fig:ushour}) since many people tend to post photographs in real time. We also match different photographs showing the same activity or concept by content analysis. Therefore our approach is more advantageous for understanding global spreads of cultural preferences than methods based on shared links or urls.

Secondly, the cultural lifestyles in user photographs tend to be more similar between friend pairs than non-friends pairs, a phenomenon known as social correlation~\cite{anagnostopoulos2008influence}. We are not aware of any previous work which attempts to measure the user similarity between tied individuals in social media by automatically analyzing \textit{visual content}. Although some studies have examined how Facebook profile photographs are correlated with user attributes such as race \cite{huang2013cultural} or gender \cite{hum2011picture}, these works did not investigate the role of network ties or any individual level's diffusion.

One important question is whether such a correlation is an outcome of social interaction (influence or induction) or simply an artifact due to homophily \cite{anagnostopoulos2008influence,sharma2016distinguishing}. Although we are unable to fully verify the causal relationship in our observational study, we found several evidences and indicators showing the possibility of an effect of social influence.

Our result is in contrast to the study by Sharma and Cosley (2016) who have proposed and applied the PME test on data from Flickr, Flixster, Last.fm, and Goodreads, and reported that the effect of influence is very small. By applying the same method, we found the effect of predictive influence is significant. We conjecture that this inconsistency could be caused by the fact that Facebook is a more \textbf{friendship-oriented} medium (\ie, most content come from friends) than the others studied in \cite{sharma2016distinguishing}.

\subsection{Limits and significance of the current study}

There are limitations in our paper. Firstly, we assume that an user would be exposed to each photograph of each of his/her friends with an identical and fixed probability. However, there exist several factors (\eg, the number of comments) that can govern the post recommendation, which might signify the effects of more popular photographs. This was also considered as a potential weakness in the original PTE test~\cite{sharma2016distinguishing}. We leave it to the future work to investigate the effects of more ``popular'' content, which may have a higher chance of correlation or influence. 

Secondly, we did not consider non-visual cues (\eg, text or urls), which may also have interesting relations to cultural diffusion. The main reasons are, as stated earlier, i) we are interested in the effect of the visual cue and its content which was not studied before and ii) photographs are global and ubiquitous while text or urls are usually language dependent and/or specific to local regions or sub-populations, which may limit the scope of study to certain local cultures.

Finally, our analysis was based on observational data and thus we cannot rule out all possible confounding factors. This is a common issue for research where manipulations or random treatment are not possible \cite{shalizi2011homophily}. However, we controlled for user age, gender, and location and used predictive statistical tests to account for the effect of homophily. While the results are not causal, our findings are novel and significant due to following reasons: i) a \textit{correlational} link on photographic similarities between massive social media users was first shown to exist by our analysis; ii) many prior studies utilizing the same statistical test~\cite{anagnostopoulos2008influence,sharma2016distinguishing} did not find the effect of influence (vs. homophily) in social networks, however, we found the effect using a novel cue. That effect was not found in a causal relation but obtained after eliminating correlations from shared user attributes and homophily. Therefore our findings strongly demonstrate the saliency of visual cues which have been overlooked in prior studies. 

\textbf{Reproducibility:} 
We provided the exact procedures for our analysis. A code to train the same model that we used is publicly available and we elaborated our modifications in full detail. While our method is reproducible, the data is not publicly available. Nevertheless, our study and procedure can apply to any social media with visual content shared by users in a social network.

\fontsize{9.5pt}{10.5pt}
\selectfont

\bibliography{main}
\bibliographystyle{aaai}
\end{document}